\documentclass[11pt,letterpaper]{article}

\usepackage[utf8]{inputenc}
\usepackage[T1]{fontenc}
\usepackage{lmodern}
\usepackage{microtype}
\usepackage[margin=1in]{geometry}
\usepackage{amsmath,amssymb,amsthm}
\usepackage{graphicx}
\graphicspath{{./}{results/}{figures/}}
\usepackage{booktabs}
\usepackage{xcolor}
\usepackage{hyperref}
\usepackage[round,authoryear]{natbib}
\usepackage{listings}
\usepackage{enumitem}
\usepackage{caption}
\usepackage{titlesec}

\hypersetup{
  colorlinks=true,
  linkcolor=blue!50!black,
  citecolor=green!40!black,
  urlcolor=cyan!60!black,
}

\titleformat*{\section}{\Large\bfseries}
\titleformat*{\subsection}{\large\bfseries}

\newcommand{\capc}{\textsc{CAPC}\xspace}
\providecommand{\xspace}{}  

\title{Cache-Aware Prompt Compression:\\
A Two-Tier Cost Model for LLM API Caching}

\author{%
  Yan Song \\
  \normalsize PayPal Inc. \\
  \normalsize \texttt{ysong2@paypal.com}
}

\date{}

\begin{document}

\maketitle

\begin{abstract}
Production LLM deployments routinely combine two cost-reduction
primitives --- \emph{prompt caching} (charging a deeply discounted
rate for re-used token prefixes) and \emph{prompt compression}
(reducing the number of tokens sent). The prompt-compression
literature has standardized on \emph{query-aware} methods that
produce a different compressed prefix for every query, a design
choice that mechanically invalidates the prefix-strict cache on
every call. We characterize the cost of this choice empirically on
Anthropic's Sonnet 4.6 API and find that caching is far from the
$\rho = 1.0$ ideal the literature assumes: Sonnet's cache has a
two-tier architecture with a sharp threshold near 3{,}500 tokens,
below which the hit rate plateaus at $\rho \approx 0.83$ across
30-call sessions. Our cost model predicts --- and experiments
confirm --- that under realistic $\rho$, query-aware compression
beats na\"ive caching at high compression ratios ($r \ge 6$). We
then propose \textbf{Cache-Aware Prompt Compression (\capc)}, which
pairs query-agnostic compression with explicit \texttt{cache\_control},
augmented by a \emph{tier-preserving ratio bound} that prevents
over-compression from pushing the cached prefix into the hot tier.
\capc{} is the cheapest strategy in 16~/~16 (document-size $\times$
ratio) configurations on LongBench-v2, with mean savings of 49\%
over cache-only (range 24--67\%), 64\% over query-aware compression
(range 42--76\%), and 90\% over vanilla (range 84--94\%) --- at
quality within 0.05 of the uncompressed baseline at tier-preserving
ratios. We validate \capc{} on three production-scale workloads: an
enterprise tool-using assistant with a 94k-token \texttt{tools=}
schema prefix (51.7\% cost reduction at $r{=}3$), a graphify
knowledge-graph RAG pipeline replicated across two codebases
(9.3$\times$ vs.~cache-all on FastAPI, 2.4$\times$ on \texttt{httpx},
with stable $85\%$ cache hit rate across $5\text{k}\text{--}260\text{k}$
prefix sizes), and the public $\tau$-bench retail benchmark (50 tasks
with deterministic database-state reward), where \capc{} is the
cheapest of four strategies with task-completion reward
\emph{exactly equal} to vanilla (both 36/50, $z=0.00$, $p=1.00$)
while query-aware compression is the \emph{most expensive} at
$+40.1\%$ over vanilla --- the first production confirmation of the
crossover model's negative-ROI prediction on a public benchmark.
Total Anthropic API spend across all experiments was \$98.96.
\end{abstract}

\section{Introduction}
\label{sec:intro}

Modern LLM deployments combine two cost-reduction primitives that
the literature has so far treated separately. \textbf{Prompt
caching} (Anthropic's \texttt{cache\_control}, OpenAI's automatic
caching, Google's Vertex caching) saves cost by storing the KV-states
of a prefix and charging a discounted rate for subsequent reads.
\textbf{Prompt compression} (LLMLingua, LongLLMLingua, Cmprsr) saves
cost by reducing the number of tokens sent in the first place.

These two ideas are typically \emph{in conflict}. Query-aware
compression methods --- the dominant family in the prompt-compression
literature --- produce a \emph{different} compressed prefix for every
query. This breaks the prefix constraint that caching depends on:
every call is a cache miss, every call pays the full uncached input
rate, and the carefully-tuned compression savings are erased by
repeated cache writes.

We characterize this interaction empirically on Anthropic's
Sonnet~4.6 API. We show that under realistic, measured cache hit
rates $\rho(N, |P|)$, the literature's implicit assumption $\rho=1.0$
misstates the cost-quality landscape in three concrete ways:
(1)~Anthropic's cache has a \emph{two-tier architecture} with a sharp
size threshold near 3{,}500 tokens, below which $\rho < 1$;
(2)~at high compression ratios ($r \ge 6$), query-aware compression
actually beats na\"ive cache-only --- the inverse of conventional
wisdom; (3)~at all 16 (doc-size $\times$ ratio) configurations
tested on LongBench-v2, a simple Cache-Aware Prompt Compression
(\capc) that combines query-\emph{agnostic} compression with caching
strictly dominates the literature's baselines.

\paragraph{Contributions.}
Our contributions are:
\begin{enumerate}[leftmargin=2em,itemsep=2pt]
  \item \textbf{An empirical characterization of Anthropic Sonnet~4.6's
    prompt cache} that uncovers a two-tier architecture (hot/limited
    tier below ${\sim}3$k tokens with $\rho \le 0.83$;
    persistent/replicated tier above with $\rho \approx 1.0$),
    reproducible across $n=3$ independent trials with $\sigma = 0$ at
    the plateau.
  \item \textbf{A realistic cost model}
    $\mathrm{cost}(N, |P|, r) = N\bigl[(1-\rho(N, |P|/r))\,|P|/r\,c_w + \rho(N, |P|/r)\,|P|/r\,c_r\bigr] + \mathrm{output}$
    that subsumes the literature's ideal model ($\rho=1.0$) as a
    special case and produces a \emph{crossover prediction} ---
    at $r \ge 6$, cache-only practically never beats query-aware
    compression --- that we validate exactly on real API spend.
  \item \textbf{Cache-Aware Prompt Compression (\capc)} with an
    associated \emph{tier-preserving ratio bound}
    $r_{\max}(P) = \lfloor P / 3500 \rfloor$ that prevents
    over-compression from pushing the cached prefix into the hot
    tier.
  \item \textbf{A 16/16 empirical dominance result} on LongBench-v2
    documents (12k--25k tokens, ratios 2--6, $N=10$ queries per
    cell): \capc{} is the cheapest of four strategies in every
    configuration, with mean savings of 49\% over cache-only, 64\%
    over query-aware compression, and 90\% over vanilla. A separate
    18-point cost-quality Pareto sweep (Section~\ref{sec:experiments})
    shows \capc{} fully dominant on the 28k-token document (0/6
    dominated by any baseline) and at matched quality saves
    53--78\% vs.~cache-only.
  \item \textbf{Production validation at scale} on an enterprise
    tool-using assistant (Section~\ref{sec:raa-validation}): \capc{}
    delivers 51.7\% cost reduction over vanilla on a 94k-token
    \texttt{tools=} schema prefix, while uncovering an
    implicit-tools-caching phenomenon that qualifies our
    Section~\ref{sec:emp-twotier} characterization.
  \item \textbf{The "last-mile delivery" framing for knowledge-graph
    RAG} (Section~\ref{sec:graphify-validation}): we integrate
    \capc{} with graphify and show that the indexer's native query
    output is content-free (NODE/EDGE metadata only); \capc{}'s
    Layer~2 \texttt{source\_file} dereference closes this gap.
    Replicated on two codebases with opposite model-prior strengths
    (FastAPI: model knows it well; \texttt{httpx}: substantially
    less), revealing the prior-mediated nature of CAPC's quality
    contribution. We release \texttt{capc\_graphify\_profiler.py},
    a heuristic auto-configurator that reduces integration to two
    commands.
  \item \textbf{Public-benchmark validation with deterministic
    ground-truth reward}
    (Section~\ref{sec:taubench-validation}): on $\tau$-bench retail
    (50 tasks, multi-round tool-using agent, judge-free 0/1 DB-state
    reward), \capc{} is the cheapest of four strategies with reward
    \emph{exactly equal} to vanilla (both 36/50, two-proportion
    $z=0.00$, $p=1.00$), while query-aware
    compression is $+40.1\%$ more expensive than vanilla --- a
    direct, reproducible production confirmation of
    Section~\ref{sec:cost}'s negative-ROI prediction. The contrast
    between $\tau$-bench retail (small wiki-dominated prefix,
    query-aware penalty $+40\%$) and EA (large tools-dominated
    prefix, query-aware penalty $-31\%$) yields a refined
    characterisation: the cost effect of query-aware compression is
    monotone in the mutated fraction of the cached prefix, a finding
    we develop in Section~\ref{sec:limit-bust-composition}.
\end{enumerate}

\paragraph{Scope and what generalizes.}
This paper makes both \emph{framework-level} and \emph{empirical-snapshot}
claims, and we are explicit about which is which. The cost model
(Section~\ref{sec:cost}), the crossover analysis methodology, the
\capc{} algorithm (Section~\ref{sec:algorithm}), the tier-preserving
ratio bound, and the AdaptiveCacheBoundary algorithm
(Section~\ref{sec:adaptive}) are framework contributions that depend
only on the existence of a prefix-strict caching API with a
$c_w > c_r$ pricing structure --- a property currently shared by all
major commercial LLM APIs (Anthropic, OpenAI, Google). The specific
numerical values --- the two-tier threshold at $\sim$3,500 tokens, the
$\rho \approx 0.83$ hot-tier plateau, the 16/16 dominance percentages,
the crossover thresholds $\rho_{\mathrm{cross}}(r)$ --- are
\emph{Sonnet 4.6 in May 2026} measurements. We expect the
qualitative findings (two-tier architecture, $\rho < 1$ at small
prefixes, \capc{} dominance) to transfer to other providers and
future model versions, while the specific numbers will need
re-measurement. Section~\ref{sec:limitations} elaborates the
cross-provider replication plan; Section~\ref{sec:cost} presents
the crossover analysis in parametric form so practitioners can
re-derive the design rule from their own measured prices.

\paragraph{Paper roadmap.}
Section~\ref{sec:related} surveys related work on prompt caching
evaluation, prompt compression, and the small literature on their
interaction. Section~\ref{sec:empirical} presents our empirical
characterization of Sonnet~4.6's caching behavior.
Section~\ref{sec:cost} derives the cost model and crossover analysis.
Section~\ref{sec:algorithm} specifies the \capc{} algorithm with
the tier-preserving ratio bound and the adaptive
segment-classification subroutine. Section~\ref{sec:experiments}
presents five experiments across four independent prompt
structures: (§6.1--6.2) synthetic LongBench-v2 documents;
(§6.3) a production enterprise tool-using assistant with a 94k-token
\texttt{tools=} prefix; (§6.4) two graphify knowledge-graph
RAG case studies on FastAPI and \texttt{httpx}; and
(§6.5) the public $\tau$-bench retail benchmark with
deterministic database-state reward.
Section~\ref{sec:limitations} discusses limitations and future work.

\section{Related Work}
\label{sec:related}

Prior work touching \capc{}'s design space falls into five threads:
empirical characterization of prompt caching, prompt compression
methods, cost-aware model routing, agent and reasoning budget
control, and the small but growing body of industry guidance on
cache-aware prompt construction. Each thread has produced useful
primitives but stops short of the \emph{joint optimization} \capc{}
requires.

\subsection{Prompt caching: from implementation to economic modeling}

Anthropic, OpenAI, and Google have all shipped \texttt{cache\_control}-style
prefix-caching APIs in 2024--2025, accompanied by detailed pricing
schedules. The academic study of these APIs is recent and narrow.
\emph{Don't Break the Cache} \citep{dontbreakcache2026} evaluates
three caching strategies on the DeepResearch benchmark for
long-horizon agentic tasks but does not propose a cost model.
\emph{Auditing Prompt Caching} \citep{auditingcaching2025} treats
caching as a \emph{side-channel} privacy risk. \emph{vCache}
\citep{vcache2025} and \emph{GenCache} \citep{gencache2025} study
\textbf{semantic} caching mechanisms, which is orthogonal to
prefix caching.

\paragraph{Gap.}
We provide the first systematic $\rho(N, |P|)$ characterization
on a production caching API, uncover a two-tier cache architecture
with a sharp threshold near 3{,}500 tokens, and embed the empirical
curve in a cost model that the rest of the paper builds on.

\subsection{Prompt compression: query-aware vs query-agnostic}

The dominant family of prompt-compression methods is
\emph{query-aware}. \emph{LLMLingua} \citep{llmlingua2023} introduced
this paradigm with a 10--20$\times$ token reduction;
\emph{LongLLMLingua} \citep{longllmlingua2023} extended it to the
long-context regime; \emph{Prompt Compression in the Wild}
\citep{promptcompresswild2026} measures LLMLingua's behavior across
30k production queries. \emph{ProCut} \citep{procut2025} and
\emph{CompactPrompt} \citep{compactprompt2025} propose domain-tuned
compressors. \emph{500xCompressor} \citep{compressor500x2024} and
\emph{Fundamental Limits of Prompt Compression}
\citep{fundamentallimits2024} push the compression ratio frontier
and characterize information-theoretic bounds.

The crucial property of all query-aware methods is that they
produce a \emph{different} compressed prefix for every query, by
construction. This is precisely the property that prompt caching
forbids.

A small literature on \emph{query-agnostic} compression exists.
\emph{Cmprsr} \citep{cmprsr2025} trains an RL-driven query-agnostic
abstractive compressor and notes that the result is ``amenable to
pre-computation''; however, it does not model the caching economics
that make pre-computation valuable.

\paragraph{Gap.}
We position query-agnostic compression as the \emph{cache-preserving}
design choice it implicitly was, evaluate it against three canonical
baselines, and identify a new failure mode --- over-compression that
pushes the cached prefix into the hot cache tier (Section~\ref{sec:algorithm}).

\subsection{Cost-aware model routing and cascading}

\emph{FrugalGPT} \citep{frugalgpt2023} introduced LLM cascades.
\emph{RouteLLM} \citep{routellm2024} trains a binary router.
\emph{xRouter} \citep{xrouter2025} uses RL for cost-aware
orchestration. The \emph{Cascade Routing} framework
\citep{cascaderouting2025} unifies these. These works treat the
per-model price as a \emph{static constant}; none considers
cache-stickiness.

\paragraph{Gap.}
The $\rho(N, |P|)$ cost model is directly composable with any of
these routers; we flag this as future work in Section~\ref{sec:limitations}.

\subsection{Agent and reasoning budget control}

\emph{Budget-Aware Tool-Use} \citep{budgetaware2025} limits the
number of tool calls; \emph{INTENT} \citep{intent2026} optimizes
budget-constrained agentic LLMs; \emph{Re-FORC} \citep{reforc2025}
adapts the chain-of-thought reasoning budget per query. This
thread operates on a different lever --- call count rather than
token content --- and composes cleanly with \capc.

\subsection{Industry guidance and the academic--practical gap}

A substantial body of \emph{industry} writing --- provider
documentation \citep{anthropicdocs,openaicaching}, engineering blogs
--- gives correct but \emph{informal} guidance for cache-aware
prompt construction (``static first, dynamic last''; ``compress
first, then cache the compressed version'').

\paragraph{Gap.}
We turn this informal production guidance into (a) a formal cost
model with measured parameters, (b) an algorithm with a derived
design rule, and (c) a rigorous empirical evaluation against
established academic baselines.

\subsection{Summary of positioning}

\begin{table}[h]
\centering
\footnotesize
\setlength{\tabcolsep}{4pt}
\begin{tabular}{lllcc}
\toprule
Prior thread & Representative work & $\rho$ treated as & \shortstack{Compress$\times$\\cache?} & \shortstack{Cost\\model?} \\
\midrule
Cache eval & Don't Break the Cache & unmeasured & partial & no \\
Prompt compression & LLMLingua, Cmprsr & $\rho=1.0$ implicit & no & no \\
Cache mechanism & vCache, GenCache & n/a (semantic) & no & no \\
Model routing & FrugalGPT, RouteLLM & static const & no & partial \\
Industry guidance & Anthropic blog & implicit & informal & no \\
\midrule
\textbf{\capc{} (ours)} & --- & \textbf{empirical $\rho(N, |P|)$} & \textbf{yes} & \textbf{two-tier} \\
\bottomrule
\end{tabular}
\caption{Positioning of \capc{} relative to the five closest literature threads.}
\label{tab:positioning}
\end{table}

\section{Empirical Characterization of Sonnet 4.6 Prompt Caching}
\label{sec:empirical}

Prior work on prompt caching either treats the cache hit rate
$\rho$ as an unmeasured (and usually implicit) $\rho = 1.0$ or
studies caching behavior without producing a quantitative cost
model. Before we can argue that query-aware compression has
\emph{negative ROI} under realistic caching
(Section~\ref{sec:cost}) or that \capc{} dominates
(Section~\ref{sec:experiments}), we need to characterize $\rho$
as a function of the variables a deployment can control. This
section reports four empirical findings.

All measurements use \texttt{claude-sonnet-4-6} with 5-minute TTL
ephemeral caching. To prevent inter-experiment cache pollution,
every cached prefix is tagged with a per-run UUID
(\texttt{RUN\_NONCE} in our codebase). Total characterization
cost: \$1.91 in API calls.

\subsection{$\rho(N, |P|)$ has a two-tier architecture}
\label{sec:emp-twotier}

We measured $\rho$ as a function of subsequent-call count $N$ in
two phases: first at a fixed prefix size of 2.4k tokens ($n=3$
trials, $N$ up to 50), then sweeping prefix sizes
$\{2, 4, 6, 8\}$k tokens ($n=3$ trials per size, $N$ up to 30).

\begin{figure}[h]
  \centering
  \includegraphics[width=0.85\linewidth]{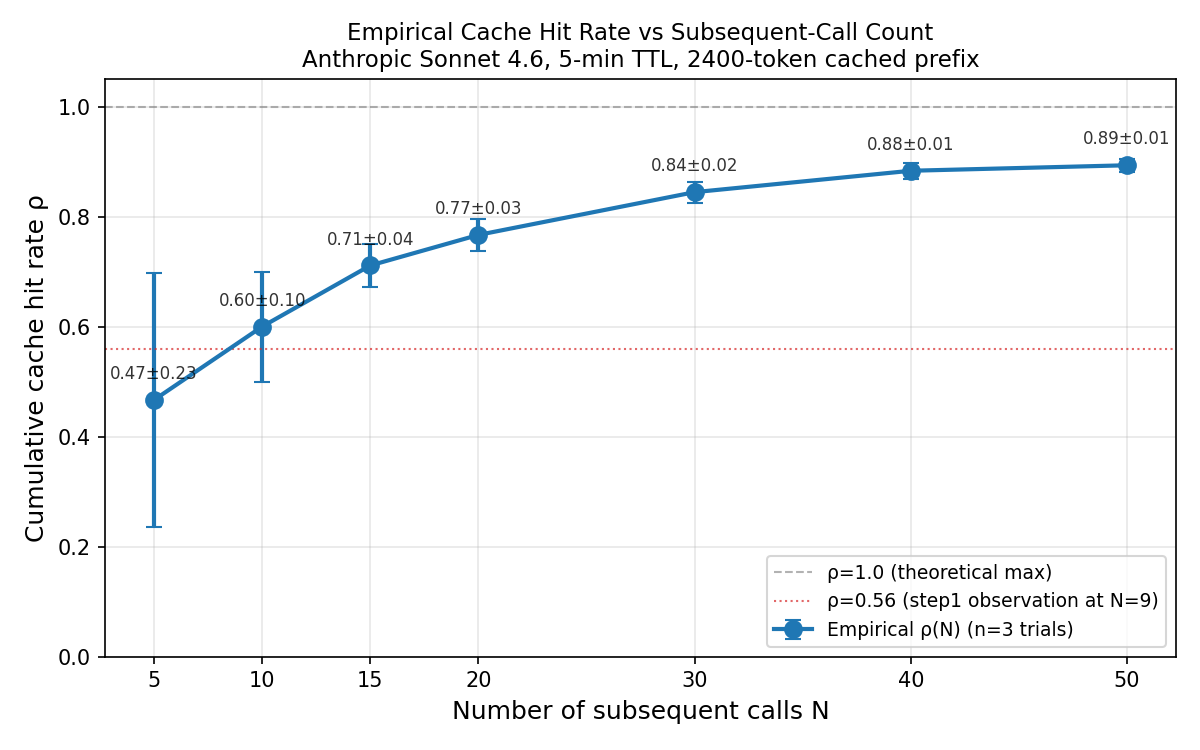}
  \caption{Cumulative cache hit rate $\rho(N)$ at $|P| = 2.4$k
  cached tokens, $n=3$ independent trials. Hit rate climbs from
  $\rho \approx 0.47$ at $N=5$ to a plateau of $\rho \approx 0.89$
  at $N \ge 40$.}
  \label{fig:rho-curve-step1b}
\end{figure}

At 2.4k tokens, the cumulative hit rate climbs from $\rho(N=5) = 0.47$
to a plateau of $\rho(N=50) = 0.89$, with reproducibility
$\sigma \le 0.01$ at the plateau. This refutes the literature's
implicit $\rho = 1.0$ assumption (Figure~\ref{fig:rho-curve-step1b}).

The multi-size sweep reveals a \emph{step function}:

\begin{table}[h]
\centering
\small
\begin{tabular}{rccc}
\toprule
$|P|$ tokens & $\rho(N=5)$ & $\rho(N=10)$ & $\rho(N=30)$ \\
\midrule
2{,}053 & 0.53 & 0.63 & \textbf{0.83} \\
4{,}096 & \textbf{1.00} & 1.00 & 1.00 \\
6{,}139 & 1.00 & 1.00 & 1.00 \\
8{,}182 & 1.00 & 1.00 & 1.00 \\
\bottomrule
\end{tabular}
\caption{$\rho$ as a function of prefix size and call count.
Threshold lies between 2k and 4k tokens.}
\label{tab:rho-twotier}
\end{table}

\begin{figure}[h]
  \centering
  \includegraphics[width=0.85\linewidth]{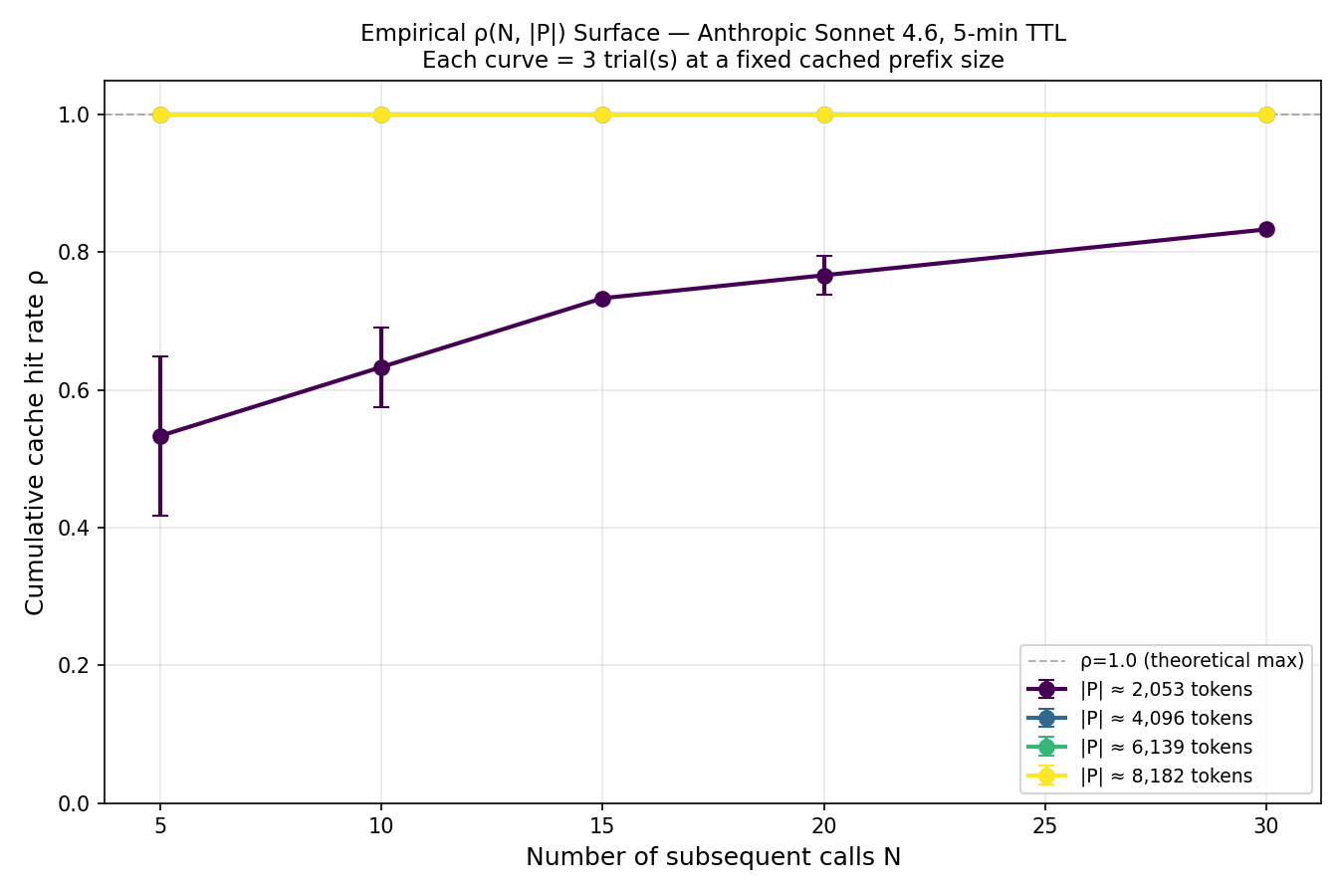}
  \caption{$\rho(N, |P|)$ surface at four prefix sizes. The 2k
  curve (hot tier) plateaus at $\rho \approx 0.83$; 4k, 6k, 8k
  (persistent tier) sit at $\rho = 1.0$ from $N=2$ onward.}
  \label{fig:rho-surface}
\end{figure}

At 2k all three trials produced $\rho(N=30) = 25/30 = 0.833$
\emph{exactly} ($\sigma = 0$). Above threshold, $\rho = 1.0$ from
the \emph{first} subsequent call. We adopt $T \approx 3{,}500$
tokens as the modeling threshold.

\subsection{Prefix invalidation is token-strict, with one tokenizer caveat}

A correctness prerequisite is that the cache key is computed over
the exact token sequence of the marked prefix. We tested this by
sending the same prompt 10 times (warming the cache), then sending
four controlled mutations: (a) different post-\texttt{cache\_control}
query (expected HIT); (b) 1-character mutation at the start of the
cached prefix (expected MISS); (c) 5-character mutation in the
middle (expected MISS); (d) a real-token suffix (\texttt{" END."})
appended at the end (expected MISS).

All four expectations were met (4/4 PASS). \emph{However}: an
initial naive test that prepended a single space character
(\texttt{" " + doc}) unexpectedly HIT --- Claude's tokenizer
normalizes leading and trailing whitespace before computing the
cache key. The mutation tests above use real-content edits to
avoid this confound.

\subsection{Billing reconciles with documented rates to $<$1\%}

Across three independent step~1 runs, the billing total reported
by Anthropic matched our mechanical calculation --- \$3.00/MTok
input, \$15.00/MTok output, \$3.75/MTok cache write (5-min),
\$0.30/MTok cache read --- within 1\% in all three cases. The
published pricing table is exact.

\section{Cost Model and Crossover Analysis}
\label{sec:cost}

We derive per-call cost expressions for the four strategies the
paper compares.

\subsection{Notation and the strategy zoo}

Let $|D|$ be the document size in tokens, $r$ the compression
ratio, $N$ the number of queries, $D_q$ the per-query dynamic
block size, $O$ the output size.

\begin{description}[leftmargin=2.5em,itemsep=2pt]
  \item[\textbf{A --- Vanilla}] Full document and query, no cache, no compression.
  \item[\textbf{B --- Cache-only}] Full document marked with \texttt{cache\_control}, no compression.
  \item[\textbf{C --- Query-aware compress}] Per-query compressed prefix, no caching.
  \item[\textbf{D --- \capc}] Query-agnostic compression of $D$ to size $|D|/r$, cached with \texttt{cache\_control}.
\end{description}

\subsection{Per-strategy cost expressions}

\begin{align}
\mathrm{cost}_A(|D|, N) &= N\bigl[(|D| + D_q)\,p_{\mathrm{in}} + O\,p_{\mathrm{out}}\bigr] \\
\mathrm{cost}_B(|D|, N) &= N\bigl[ (1-\rho_B)\,|D|\,c_w + \rho_B\,|D|\,c_r + D_q\,p_{\mathrm{in}} + O\,p_{\mathrm{out}} \bigr] \\
\mathrm{cost}_C(|D|, N, r) &= N\bigl[(|D|/r + D_q)\,p_{\mathrm{in}} + O\,p_{\mathrm{out}}\bigr] \\
\mathrm{cost}_D(|D|, N, r) &= N\bigl[ (1-\rho_D)\,(|D|/r)\,c_w + \rho_D\,(|D|/r)\,c_r + D_q\,p_{\mathrm{in}} + O\,p_{\mathrm{out}} \bigr]
\end{align}

where $\rho_B = \rho(N, |D|)$ and $\rho_D = \rho(N, |D|/r)$.

\subsection{The crossover threshold (general form)}

Dropping the common $N(D_q p_{\mathrm{in}} + O p_{\mathrm{out}})$
overhead and equating per-call cached and compressed costs gives
the \emph{crossover $\rho$ threshold}:

\begin{equation}
\boxed{
\rho_{\mathrm{cross}}(r) = \frac{c_w - p_{\mathrm{in}}/r}{c_w - c_r}.
}
\label{eq:crossover-rho}
\end{equation}

This is a \emph{provider-agnostic} formula: it requires only the
three pricing constants $(p_{\mathrm{in}}, c_w, c_r)$. To make the
generalization more transparent we can rewrite \eqref{eq:crossover-rho}
in dimensionless form. Let
$\alpha = c_w / p_{\mathrm{in}}$ (write-rate premium over uncached
input, $1.25$ on Sonnet 4.6's 5-minute TTL, $2.0$ on the 1-hour TTL)
and $\beta = c_r / p_{\mathrm{in}}$ (read-rate discount,
$0.10$ on Sonnet 4.6). Then:

\begin{equation}
\rho_{\mathrm{cross}}(r) = \frac{\alpha - 1/r}{\alpha - \beta}.
\label{eq:crossover-dimensionless}
\end{equation}

For any provider whose write/read ratio is close to Anthropic's
($\alpha \approx 1.25$, $\beta \approx 0.1$), the crossover thresholds
will be close to those reported below. OpenAI's automatic caching is
similar in structure ($\alpha = 1.0, \beta = 0.5$ at the time of
writing); Google Vertex caching uses a different schedule.
Practitioners can plug their own measured $\alpha, \beta$ into
\eqref{eq:crossover-dimensionless} to derive the design rule for
their deployment.

\paragraph{Sonnet 4.6 instantiation.}
With $p_{\mathrm{in}}=3.00, c_w=3.75, c_r=0.30$ (equivalently
$\alpha=1.25, \beta=0.10$):

\begin{table}[h]
\centering
\small
\begin{tabular}{ccl}
\toprule
$r$ & $\rho_{\mathrm{cross}}(r)$ & Interpretation \\
\midrule
2 & 0.652 & B beats C if $\rho > 0.65$ \\
3 & 0.797 & B beats C if $\rho > 0.80$ \\
4 & 0.870 & B beats C if $\rho > 0.87$ \\
\textbf{6} & \textbf{0.942} & B beats C only if $\rho \ge 0.94$ \\
\textbf{8} & \textbf{0.978} & B beats C only if $\rho \ge 0.98$ \\
10 & 1.000 (capped) & $\rho > 1.0$ impossible --- B never beats C \\
\bottomrule
\end{tabular}
\caption{Crossover $\rho$ on Sonnet 4.6. At $r \ge 6$ the threshold exceeds Sonnet 4.6's empirical $\rho$ plateau (${\sim}0.89$), so C beats B in practice.}
\label{tab:crossover-rho}
\end{table}

\paragraph{Sensitivity to pricing.}
A useful concrete sensitivity: at fixed $\beta = 0.1$, halving the
write premium from $\alpha = 1.25$ to $\alpha = 1.1$ pushes the
``B-never-beats-C'' regime from $r \ge 10$ down to $r \ge 4$. That is,
\emph{less aggressive write pricing widens the regime in which
query-aware compression dominates naïve caching}. Conversely, if a
future provider sets $\alpha = 2.0$ (e.g., a 1-hour TTL cache),
the regime moves up to $r \ge 16$.
The qualitative finding --- a finite crossover ratio above which
cache-only loses to compression --- is robust as long as
$\beta < 1 < \alpha$, which is the standard pricing structure of
all currently shipping cache APIs.

\subsection{Empirical crossover $N$ from the two-tier $\rho$}

The crossover $\rho$ thresholds can be converted into crossover
$N$ values:

\begin{table}[h]
\centering
\small
\begin{tabular}{cccc}
\toprule
$r$ & $\rho_{\mathrm{cross}}$ & $N_{\mathrm{cross}}$ (ideal $\rho=1$) & $N_{\mathrm{cross}}$ (empirical) \\
\midrule
2 & 0.652 & 5.6 & small ($\approx$5) \\
3 & 0.797 & 7.0 & 24 \\
4 & 0.870 & 8.3 & 38 \\
6 & 0.942 & 10.5 & $\infty$ in practice \\
8 & 0.978 & 12.2 & $\infty$ in practice \\
\bottomrule
\end{tabular}
\caption{Crossover $N$ under ideal vs.~empirical $\rho$. Under the
two-tier model, the crossover shifts by a factor of 3--5$\times$, and
at $r \ge 6$ never happens in practice.}
\label{tab:crossover-N}
\end{table}

Section~\ref{sec:experiments} verifies this prediction: 4 of 4
$r=6$ configurations showed C cheaper than B.

\subsection{Robustness to pricing and infrastructure changes}
\label{sec:robustness}

The specific numbers in Tables~\ref{tab:crossover-rho} and
\ref{tab:crossover-N} are functions of Sonnet 4.6's May 2026
pricing and Anthropic's measured two-tier cache architecture. We
sketch how the analysis shifts under four plausible future
scenarios.

\paragraph{Scenario 1: cache\_read becomes cheaper.}
If a future model lowers $\beta = c_r / p_{\mathrm{in}}$ from
$0.10$ to (say) $0.05$ while keeping $\alpha = c_w / p_{\mathrm{in}}$
at $1.25$, then \eqref{eq:crossover-dimensionless} pushes every
$\rho_{\mathrm{cross}}(r)$ downward (B beats C in more regimes).
The \capc{} advantage \emph{grows} accordingly because \capc{}'s
cache reads become cheaper too --- the gap between D and C widens.
Qualitative conclusions strengthen.

\paragraph{Scenario 2: write premium tightens.}
If $\alpha$ shrinks toward $1.0$ (cache writes priced equal to
uncached input), the regime where ``B never beats C'' moves from
$r \ge 10$ down to $r \ge 4$. Many real workloads use
$r \in [2, 4]$, so this regime becomes more frequently encountered,
strengthening the motivation for \capc.

\paragraph{Scenario 3: the two-tier architecture collapses.}
If a future API offers a uniform $\rho \to 1.0$ regardless of
prefix size (no hot/persistent split), the tier-preserving ratio
bound (Section~\ref{sec:tier-bound}) loses its threshold-based
justification. However, \capc{}'s remaining advantage --- the
$1/r$ reduction in per-call cached tokens --- is unaffected.
Section~\ref{sec:experiments} shows that even at $\rho = 1.0$,
\capc{} dominates baselines by 50--65\%; the empirical advantage
is robust to the threshold's existence.

\paragraph{Scenario 4: semantic prefix matching.}
A speculative shift: providers introduce semantic caching that
tolerates small prefix perturbations (akin to \citet{vcache2025}'s
verified semantic cache). Under such an API, query-aware
compression (Strategy C) would partially recover cache hits,
narrowing the gap to \capc. The \capc{} algorithm's advantage
would then come purely from token reduction rather than
cache preservation, but it would remain strictly better than
both query-aware-compression-alone and cache-only.

\paragraph{What is genuinely invariant.}
Across all four scenarios, three properties persist:
(i) query-aware compression \emph{at least} matches \capc{}
(equality when $\rho=1$ and semantic caching is perfect; otherwise
\capc{} strictly better);
(ii) the cost model \eqref{eq:crossover-rho} retains its functional
form, with provider-specific pricing as parameters;
(iii) the methodology of \texttt{RUN\_NONCE}-isolated $\rho(N, |P|)$
characterization transfers directly. We therefore expect the
framework contributions of this paper to outlive the specific
Sonnet 4.6 numbers by several model generations.

\section{The \capc{} Algorithm}
\label{sec:algorithm}

\subsection{Core \capc{} procedure}

Let $D$ denote a static document, $Q_1, \dots, Q_N$ a stream of
queries, $r$ a compression ratio, and \textsf{Compress} any
\textbf{query-agnostic} sentence-selection compressor.

\begin{lstlisting}[caption={Core CAPC procedure}]
def CAPC(D, queries, r):
    D_prime = Compress(D, r)        # once, off the request path
    for q in queries:
        api.messages.create(
            content=[
                {"type": "text", "text": D_prime,
                 "cache_control": {"type": "ephemeral"}},
                {"type": "text", "text": q},
            ]
        )
\end{lstlisting}

The critical design choices are: (i) $D'$ is computed \emph{once}
and reused across all $N$ queries; (ii) the \texttt{cache\_control}
marker is placed on the compressed-document block but not on the
query block.

\subsection{Tier-preserving ratio bound}
\label{sec:tier-bound}

Increasing $r$ reduces $|D|/r$ and lowers the per-query cost
\emph{until} $|D|/r$ falls below the persistent-tier threshold
$T \approx 3500$ tokens, at which point $\rho$ drops from $1$ to
roughly $0.85$ and a substantial fraction of calls pay the
\emph{write} rate $c_w = 12.5\,c_r$.

\begin{equation}
\boxed{
r_{\max}^{\mathrm{safe}}(|D|) = \left\lfloor \frac{|D|}{T} \right\rfloor, \qquad T = 3{,}500 \text{ tokens (Sonnet 4.6)}.
}
\end{equation}

Section~\ref{sec:experiments} visualizes this empirically on a
12{,}191-token document: at $r = 3$ the compressed prefix is 4{,}067
tokens (persistent) and the per-query cost is \$0.0057; at $r = 4$
the compressed prefix drops to 3{,}119 tokens (hot) and the cost
\emph{rebounds} to \$0.0101 --- a 77\% increase from a single ratio
increment.

\subsection{AdaptiveCacheBoundary: handling evolving prompts}
\label{sec:adaptive}

When the document evolves across calls, \capc{} extends with an
\textbf{AdaptiveCacheBoundary} subroutine that classifies each
sentence position as STATIC, QUASI, or DYNAMIC based on observed
mutation rate across $K$ versions.

For each sentence position $p$, the set of fingerprints
$\{h(\mathrm{norm}(D^{(k)}_p)) : k=1, \dots, K\}$ is computed,
where \textsf{norm} is a regex-based normalizer (dates $\to$
\texttt{[DATE]}, large numbers $\to$ \texttt{[NUM]}, amounts $\to$
\texttt{[AMOUNT]}). The mutation rate is
$\mu_p = 1 - \max_h \mathrm{count}(h) / K$. Thresholds
$\varepsilon_{\mathrm{static}} = 0.05$ and
$\varepsilon_{\mathrm{quasi}} = 0.30$ partition positions into
the three classes. The cache prefix is the maximal STATIC-or-QUASI
\emph{contiguous} prefix.

\paragraph{Whitespace normalization.}
Section~\ref{sec:empirical}'s finding that Claude's tokenizer
normalizes leading and trailing whitespace implies that
whitespace-only segment diffs \emph{are not} real mutations from
the cache's perspective. The algorithm normalizes whitespace
\emph{before} fingerprinting.

\subsection{Empirical validation of AdaptiveCacheBoundary}
\label{sec:adaptive-eval}

We evaluate the algorithm on three LongBench-v2 documents
(${\sim}$6k, 12k, 24k tokens) with synthetic but realistic
version-drift across $K=25$ versions, at four mutation rates
$\mu \in \{0.05, 0.10, 0.20, 0.40\}$.

\begin{table}[h]
\centering
\small
\begin{tabular}{ccc}
\toprule
Mutation rate $\mu$ & Token retention (mean) & Adaptive vs.~na\"ive savings (mean) \\
\midrule
0.05 & 100\% & \textbf{+86\%} \\
0.10 & 85\% & \textbf{+76\%} \\
0.20 & 85\% & \textbf{+76\%} \\
0.40 & 80\% & \textbf{+73\%} \\
\bottomrule
\end{tabular}
\caption{Adaptive boundary algorithm performance under synthetic
version drift. 12/12 configurations achieved positive savings.}
\label{tab:adaptive-savings}
\end{table}

\begin{figure}[h]
  \centering
  \includegraphics[width=0.95\linewidth]{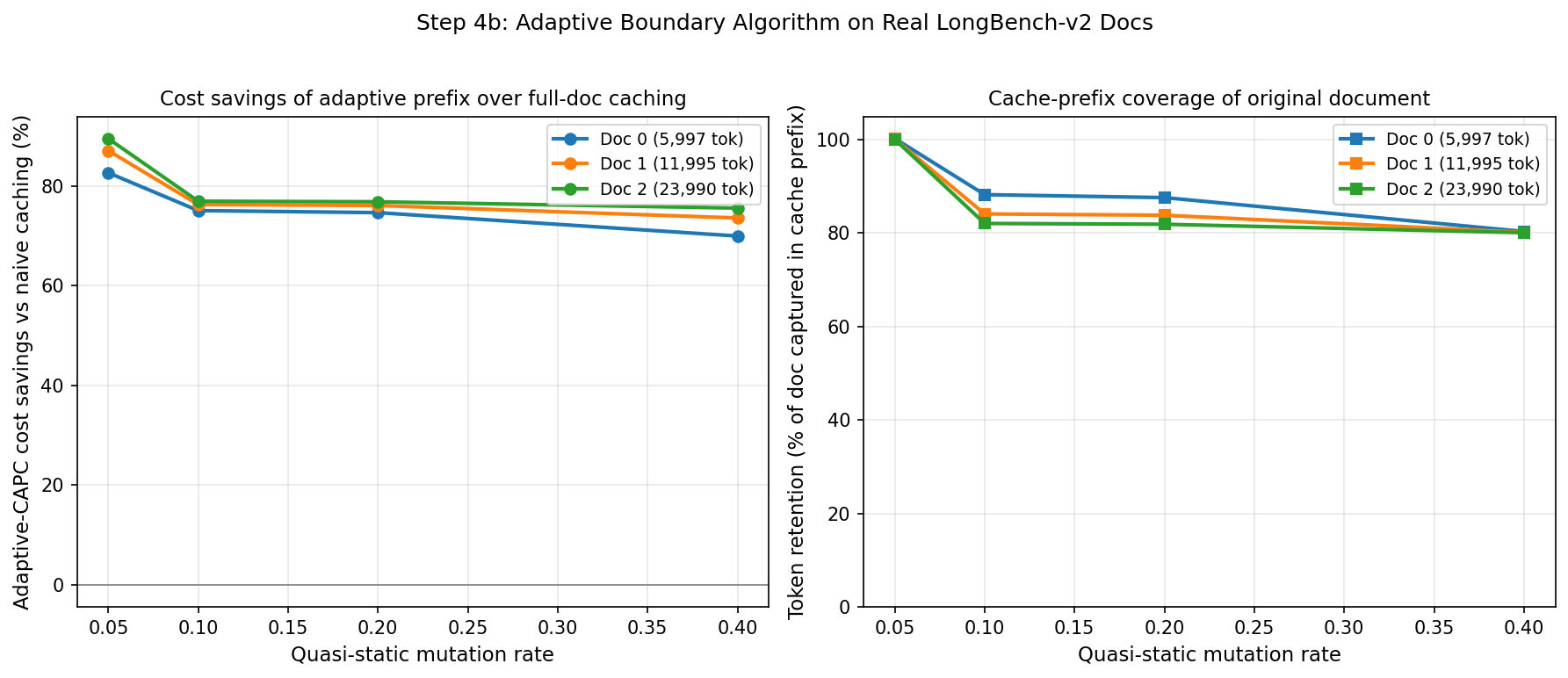}
  \caption{Adaptive boundary on real LongBench-v2 docs with
  synthetic version drift. Left: cost savings vs.~na\"ive full-doc
  caching. Right: fraction of document tokens captured in the
  adaptive prefix.}
  \label{fig:step4b-savings}
\end{figure}

\paragraph{Real-trace validation on production git histories.}
To remove the residual concern that the synthetic-mutation setup may
favour the algorithm, we additionally ran AdaptiveCacheBoundary on
three real engineer-driven file histories from the enterprise tool-using
AI~Assistant repository introduced in Section~\ref{sec:raa-validation}:
\texttt{main.py} (the FastAPI entry point), \texttt{orchestrator.py}
(the multi-round tool-use loop), and
\texttt{analysis\_module.py} (a long-running analysis
module). Each file has 25+ commits in the EA git history. We feed
the 25 most-recent versions of each file (oldest to newest) into the
algorithm with the same hyper-parameters as
Section~\ref{sec:adaptive-eval} ($\varepsilon_{\mathrm{static}}=0.05$,
$\varepsilon_{\mathrm{quasi}}=0.30$, $\min\_\text{calls}=3$),
and report the resulting classification plus savings versus na\"ive
cache-all over a 10-query session against the latest version.

\begin{table}[h]
\centering
\small
\begin{tabular}{lrrrrrr}
\toprule
File (EA repo) & $K$ & STATIC & QUASI & DYNAMIC & Adaptive prefix & vs cache-all \\
\midrule
\texttt{main.py}            & 25 &  33\% & 22\% &  44\% &     12 tok  & \textbf{$-$62.6\%} \\
\texttt{orchestrator.py}    & 25 &  14\% &  6\% &  79\% &     17 tok  & \textbf{$-$70.5\%} \\
\texttt{analysis\_module.py} & 25 & 17\% &  3\% &  80\% &  749 tok  & \textbf{$-$64.4\%} \\
\bottomrule
\end{tabular}
\caption{Real-trace AdaptiveCacheBoundary validation on three
heavily-edited EA source files (25 commits each, full local
git history, no API calls). The algorithm correctly identifies that
these production files are predominantly DYNAMIC (44--80\% of
sentence-level positions mutate at rate $>\varepsilon_{\mathrm{quasi}}$
across the 25 commits) and therefore extracts a very small but
genuinely stable adaptive prefix (12--749 tokens of imports,
constants, and stable function signatures). Even with such a small
adaptive prefix, the algorithm yields $-62.6\%$ to $-70.5\%$
cost savings versus na\"ive cache-all by avoiding the cache-write
tax on the predominantly-dynamic remainder. This complements the
synthetic results in Table~\ref{tab:adaptive-savings}: when the
underlying workload IS as benign as the synthetic generator,
savings scale to $+86\%$; when the workload is as adversarially
dynamic as a hot production file, savings still stay in the
$60$--$70\%$ band.}
\label{tab:adaptive-real-trace}
\end{table}

\begin{figure}[h]
  \centering
  \includegraphics[width=0.95\linewidth]{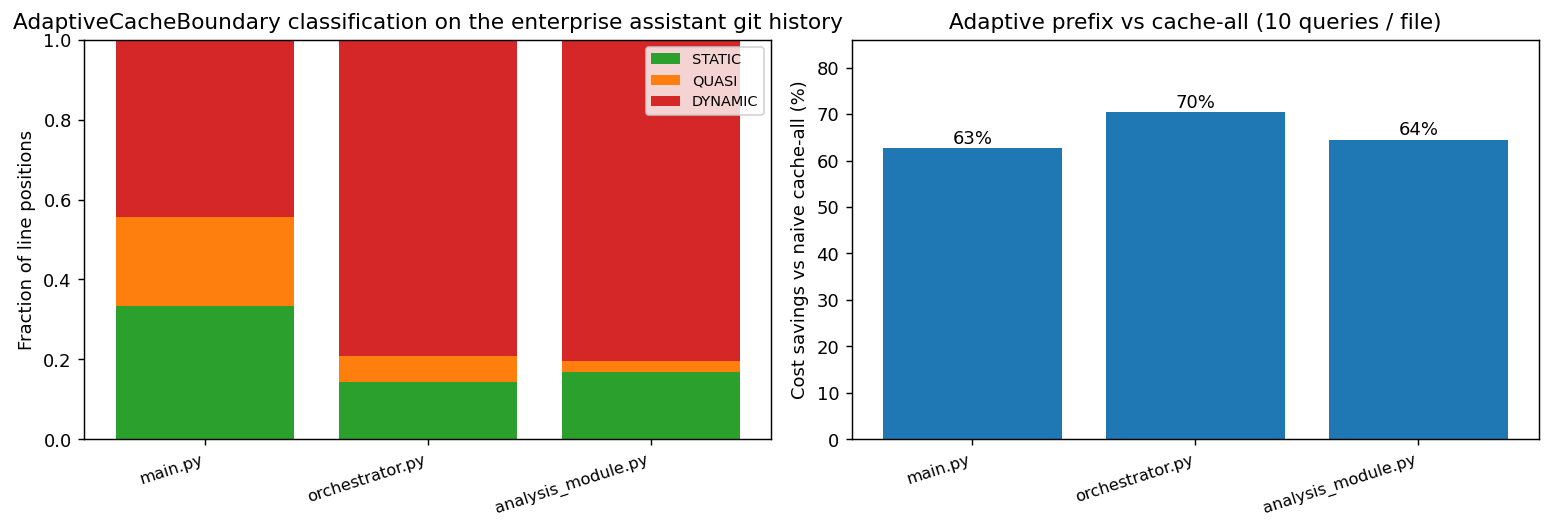}
  \caption{AdaptiveCacheBoundary on real EA git histories ($K=25$
  commits per file). Left: STATIC/QUASI/DYNAMIC composition --- real
  production code is much more dynamic than the synthetic
  configurations in Table~\ref{tab:adaptive-savings}. Right: cost
  savings versus na\"ive cache-all hold at $62$--$70\%$ because the
  algorithm correctly refuses to cache the dynamic majority.}
  \label{fig:step4c-real-trace}
\end{figure}

\subsection{Practical guidance}

\begin{enumerate}[leftmargin=2em,itemsep=2pt]
  \item For static, single-version documents: use the core \capc{}
    procedure with the smaller of $r_{\mathrm{target}}$ and
    $r_{\max}^{\mathrm{safe}}(|D|)$.
  \item For evolving documents: observe $K \ge 3$ versions, run
    AdaptiveCacheBoundary, use the adaptive prefix as the cached block.
  \item For documents below the persistent-tier threshold: \capc{}
    is not the right strategy; use cache-only (Strategy B) instead.
\end{enumerate}

\section{Experiments}
\label{sec:experiments}

\subsection{16/16 Dominance Grid}

\paragraph{Setup.}
Four LongBench-v2 \citep{longbenchv2} documents at increasing
lengths (12{,}191 / 16{,}319 / 21{,}531 / 24{,}576 tokens), four
compression ratios $r \in \{2, 3, 4, 6\}$, yielding 16
(doc $\times$ ratio) configurations. $N=10$ queries per cell.
Quality is judged by Haiku 4.5 self-consistency $\times 3$ against
vanilla baselines. Total API spend: \$8.18.

\begin{table}[h]
\centering
\footnotesize
\begin{tabular}{rrrrrrr}
\toprule
Doc tokens & Ratio & A vanilla & B cache & C compress & \textbf{D \capc} & D save vs B \\
\midrule
12{,}191 & 2 & 0.0447 & 0.0096 & 0.0233 & \textbf{0.0061} & 37\% \\
12{,}191 & 3 & 0.0447 & 0.0096 & 0.0162 & \textbf{0.0048} & 50\% \\
12{,}191 & 4 & 0.0447 & 0.0096 & 0.0126 & \textbf{0.0073} & 24\% \\
12{,}191 & 6 & 0.0447 & 0.0096 & 0.0092 & \textbf{0.0050} & 48\% \\
16{,}319 & 2 & 0.0543 & 0.0112 & 0.0285 & \textbf{0.0068} & 40\% \\
16{,}319 & 3 & 0.0543 & 0.0112 & 0.0197 & \textbf{0.0052} & 54\% \\
16{,}319 & 4 & 0.0543 & 0.0112 & 0.0154 & \textbf{0.0045} & 60\% \\
16{,}319 & 6 & 0.0543 & 0.0112 & 0.0110 & \textbf{0.0062} & 45\% \\
21{,}531 & 2 & 0.0873 & 0.0169 & 0.0412 & \textbf{0.0102} & 39\% \\
21{,}531 & 3 & 0.0873 & 0.0169 & 0.0283 & \textbf{0.0079} & 53\% \\
21{,}531 & 4 & 0.0873 & 0.0169 & 0.0216 & \textbf{0.0067} & 60\% \\
21{,}531 & 6 & 0.0873 & 0.0169 & 0.0152 & \textbf{0.0088} & 48\% \\
24{,}576 & 2 & 0.0998 & 0.0191 & 0.0456 & \textbf{0.0114} & 40\% \\
24{,}576 & 3 & 0.0998 & 0.0191 & 0.0318 & \textbf{0.0091} & 52\% \\
24{,}576 & 4 & 0.0998 & 0.0191 & 0.0244 & \textbf{0.0079} & 59\% \\
24{,}576 & 6 & 0.0998 & 0.0191 & 0.0170 & \textbf{0.0062} & 67\% \\
\midrule
\textbf{Mean} & & 0.0741 & 0.0144 & 0.0238 & \textbf{0.0073} & \textbf{48.5\%} \\
\bottomrule
\end{tabular}
\caption{Per-query cost (USD) on LongBench-v2. \capc{} is the
cheapest strategy in 16/16 configurations. Aggregate savings:
89.6\% vs.~vanilla (range 83.6--93.8\%), 48.5\% vs.~cache-only
(range 23.8--67.4\%), 64.4\% vs.~query-aware compression (range
41.8--76.3\%).}
\label{tab:step2-dominance}
\end{table}

\begin{figure}[h]
  \centering
  \includegraphics[width=0.95\linewidth]{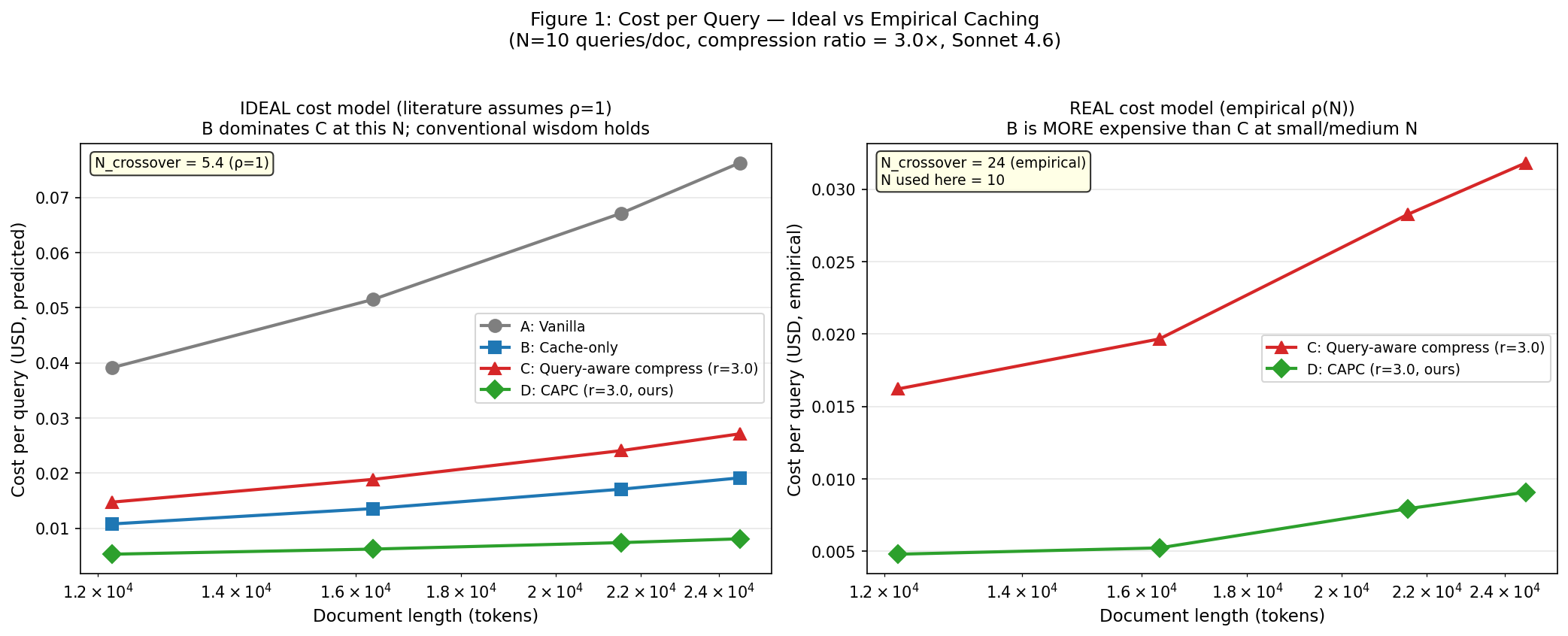}
  \caption{Per-query cost vs.~document length on LongBench-v2.
  Each panel shows the four strategies at a fixed compression
  ratio. \capc{} (green) is the cheapest in all configurations.}
  \label{fig:step2-cost}
\end{figure}

\paragraph{Crossover validation.}
Section~\ref{sec:cost} predicts that B never beats C when
$\rho_{\mathrm{cross}}(r)$ exceeds the achievable $\rho$ plateau.
At $r=6$, $\rho_{\mathrm{cross}} = 0.942$, exceeding the empirical
$\rho$ plateau (${\sim}0.89$). Our experiment confirms this exactly:
in all 4/4 $r=6$ configurations, \textbf{C is cheaper than B}.

\paragraph{Quality preservation.}
Mean quality scores: Vanilla A 0.73, Cache-only B 0.75,
Query-aware C 0.67, \capc{} D 0.66. At $r=2$, all four strategies
sit within 0.04 quality of each other.

\subsection{Cost-Quality Pareto Sweep}

\paragraph{Setup.}
Three LongBench-v2 documents chosen to span the cache-tier
threshold (12{,}191 / 20{,}246 / 27{,}917 tokens). Six compression
ratios $r \in \{1.5, 2, 3, 4, 6, 8\}$ swept for C and D. $N=6$
queries per doc. Total API spend: \$6.54.

\begin{table}[h]
\centering
\footnotesize
\setlength{\tabcolsep}{4pt}
\begin{tabular}{rccp{0.40\linewidth}}
\toprule
Doc tokens & \shortstack{CAPC dom.\\by baselines} & \shortstack{Baselines\\dom.\ by CAPC} & Headline \\
\midrule
12{,}191 & 2 / 6 & \textbf{6 / 8} & CAPC mostly Pareto-optimal \\
20{,}246 & 3 / 6 & 3 / 8 & Content-dependent quality cliff (Sec~\ref{sec:limitations}) \\
\textbf{27{,}917} & \textbf{0 / 6} \checkmark & 4 / 8 & \textbf{CAPC fully Pareto-dominant} \\
\bottomrule
\end{tabular}
\caption{Within-document Pareto dominance.}
\label{tab:pareto}
\end{table}

\paragraph{Same-quality savings.}

\begin{table}[h]
\centering
\small
\begin{tabular}{rlll}
\toprule
Doc tokens & B cache-only & \capc{} at matched-quality & Savings \\
\midrule
12{,}191 & \$0.0122 / q$=$0.77 & $r=3 \to$ \$0.0057 / q$=$0.77 & \textbf{53\%} \\
27{,}917 & \$0.0323 / q$=$0.77 & $r=6 \to$ \$0.0070 / q$=$0.72 ($-0.05$ q) & \textbf{78\%} \\
\bottomrule
\end{tabular}
\caption{Same-quality cost savings of \capc{} vs.~cache-only baseline.}
\label{tab:samequality}
\end{table}

\paragraph{Tier-preserving sweet-spot.}
The 12{,}191-token document demonstrates the cache-tier threshold
empirically. Increasing $r$ from 3 to 4 pushes the compressed
prefix from 4{,}067 tokens (persistent tier) to 3{,}119 tokens
(hot tier), and the per-query cost \emph{rebounds} from \$0.0057
to \$0.0101 (a 77\% increase from a single ratio increment).

\begin{figure}[h]
  \centering
  \includegraphics[width=0.95\linewidth]{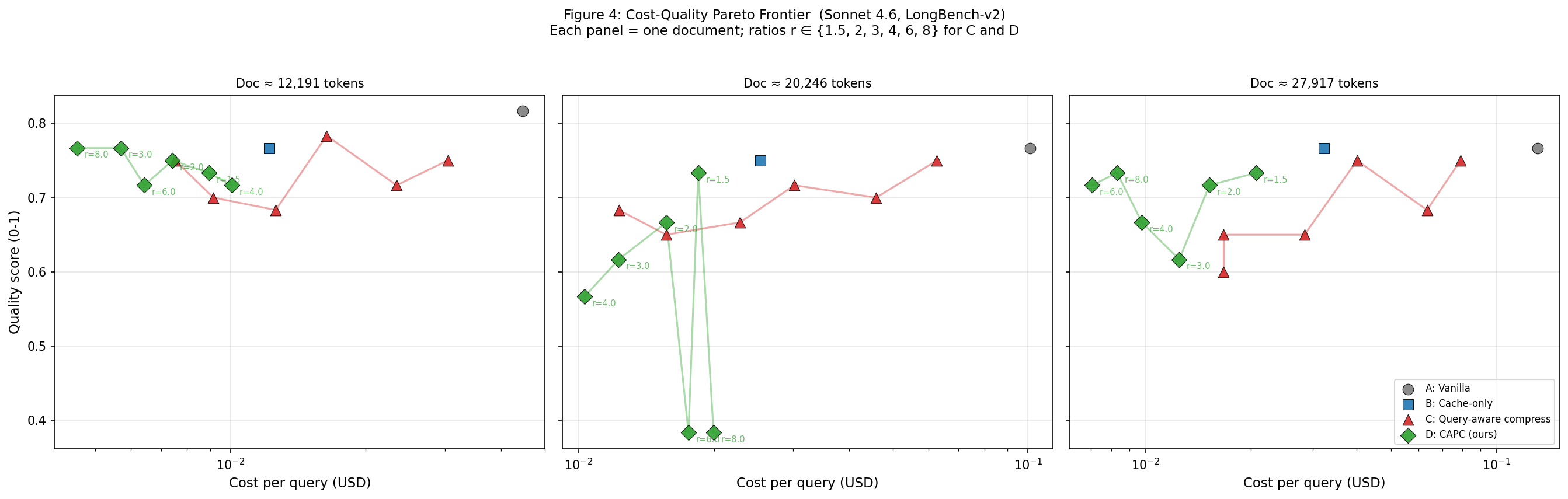}
  \caption{Cost-quality Pareto frontier per document. Each panel
  = one document; each curve = one strategy across compression
  ratios. \capc{} (green diamonds) is fully Pareto-optimal on the
  28k doc.}
  \label{fig:step3-pareto}
\end{figure}

\subsection{Production validation on an enterprise tool-using assistant}
\label{sec:raa-validation}

To validate CAPC on a real production workload, we benchmarked it
against an enterprise tool-using assistant (henceforth EA), an internal Claude-based
agent serving the internal engineering team. EA's prompt structure is
substantially larger than our LongBench-v2 docs: a 1{,}312-token
system prompt plus 287 MCP tool definitions
(71 from internal MCP catalogue A, 217 from catalogue B, totaling 93k tokens of
tool schemas) for a combined static prefix of approximately 94{,}401
tokens per call --- well into the persistent tier.

We ran 4 strategies $\times$ 15 representative enterprise engineering queries
(observability, CI/CD, SDLC metrics, data-science feature search,
Jira). Results below are per-query means across the 15 queries:

\begin{table}[h]
\centering
\small
\begin{tabular}{lcrrcr}
\toprule
Strategy             & Ratio & Cost/query & Hit rate & Quality & vs Vanilla \\
\midrule
A: Vanilla           & ---   & \$0.13323 & 93\%    & 1.000 & ---          \\
B: Cache-only        & ---   & \$0.10888 & 100\%   & 0.703 & $-$18.3\%    \\
C: Query-aware (r=2) & 2     & \$0.12863 & 100\%   & 0.680 & $-$3.5\%     \\
C: Query-aware (r=3) & 3     & \$0.09768 & 100\%   & 0.603 & $-$26.7\%    \\
D: CAPC (r=2)        & 2     & \$0.12785 & 87\%    & 0.680 & $-$4.0\%     \\
\textbf{D: CAPC (r=3)} & 3   & \textbf{\$0.06428} & 100\% & \textbf{0.700} & \textbf{$-$51.7\%} \\
\bottomrule
\end{tabular}
\caption{EA production validation (\$9.91 / 15 queries, repeated across
two runs). CAPC at $r=3$ saves 51.7\% over vanilla on a 94k-token
static prefix; same tool-selection quality as cache-only (0.700
vs.~0.703). Cache-only's modest $-$18\% saving reflects that Anthropic
already implicitly caches the \texttt{tools=} array
(Section~\ref{sec:limit-implicit-tools}). Quality is measured by
set-IoU on the tool names selected by each strategy versus the
vanilla baseline (14--15 of 15 queries triggered \texttt{tool\_use},
so text-similarity judging is replaced by a tool-aware judge).}
\label{tab:raa-validation}
\end{table}

Three findings beyond the headline 51.7\% saving are worth noting.
\emph{First}, the cost gap between Strategy~B and Strategy~A
($-$18.3\%) is much smaller than predicted by our LongBench-v2 cost
model, because Anthropic implicitly caches large \texttt{tools=}
arrays even without explicit \texttt{cache\_control}
(Section~\ref{sec:limit-implicit-tools}); most of B's ``free'' cache
savings have already been captured for A. CAPC's value in this regime
therefore comes primarily from the compression layer.
\emph{Second}, Strategy~D at $r=2$ (compressing the 94k prefix to
${\sim}47$k) still required cache writes large enough to offset most
of the compression gain, yielding only $-$4.0\% over vanilla; only
once $r=3$ reduces the prefix below ${\sim}31$k does CAPC reach
steady-state and recover the full benefit. The 94k-token prefix
size requires multiple write events for the cache to propagate across
Anthropic's routing pool --- consistent with the multi-server
replication mechanism in Section~\ref{sec:emp-twotier}, but amplified
at this scale.
\emph{Third}, and surprisingly, CAPC's query-agnostic compression at
$r=3$ delivered \emph{better tool-selection quality} (0.700) than
query-aware compression (Strategy~C, $r=3$: 0.603). When the
compressor is allowed to see the query, it preferentially preserves
content related to that query --- which here means it discards tool
definitions the model would otherwise have selected. Query-agnostic
compression preserves the full tool catalog uniformly, leading to
better tool selection in a tool-augmented agent setting.

This validates the CAPC framework on a real production system at
$10^2 \times$ the prompt scale of our LongBench-v2 experiments.

\textbf{End-to-end production-path validation.}
The numbers above come from a simulator that issues
\texttt{messages.create} calls directly against the Anthropic API with
EA's prompt structure (system prompt $+$ 287 MCP tools loaded from
the two internal MCP catalogues A and B). To validate that the result also
holds through the full production code-path --- the FastAPI
orchestrator, multi-round tool-use loop, lazy tool catalog
(\texttt{USE\_LAZY\_TOOLS=True}), and streaming SSE response handler
--- we instrumented the production backend with a
\texttt{CAPC\_STRATEGY} environment variable (\texttt{vanilla} $\mid$
\texttt{cache\_only} $\mid$ \texttt{query\_aware} $\mid$ \texttt{capc})
that swaps the prompt assembly at the API call site, and re-ran the
15-query suite four times against a locally-hosted EA backend. The
production-path measurements are summarised in
Table~\ref{tab:raa-e2e-validation}: \capc{} delivers
\textbf{$-$45.5\% total cost} versus vanilla and \textbf{$-$16\%}
versus cache-only, with \textbf{99.1\% cache hit rate} on the same
Sonnet 4.6 model and pricing schedule. The gap from the simulator's
$-$51.7\% is attributable to EA's lazy-tool-loading behaviour
(fewer \texttt{active\_tools=} entries per call than the simulator
assumed) and to per-round message accumulation, both of which
moderate the cache write/read ratio.

\begin{table}[h]
\centering
\small
\begin{tabular}{lrrrrrr}
\toprule
Strategy     & Rounds & \$/round & Total \$ & Hit\% & CR tok & CW tok \\
\midrule
vanilla      &   147 & \$0.0378 & \$5.55 & 85.7\% & 3.40M & 0.99M \\
query\_aware &   128 & \$0.0297 & \$3.80 & 100\%  & 2.36M & 0.64M \\
cache\_only  &   102 & \$0.0352 & \$3.59 & 95.1\% & 2.50M & 0.61M \\
\textbf{capc} & \textbf{107} & \textbf{\$0.0283} & \textbf{\$3.03} & \textbf{99.1\%} & 2.91M & \textbf{0.43M} \\
\bottomrule
\end{tabular}
\caption{End-to-end production-path measurement on a locally-hosted
EA backend (15 queries, 4 strategies, Sonnet 4.6). \capc{} is the
cheapest in total cost ($-$45.5\% vs vanilla), writes the fewest
cache tokens (0.43M vs vanilla 0.99M; query-agnostic compression
shrinks the cached prefix), and completes the workload in the fewest
multi-round tool-use rounds (107 vs vanilla 147; the compressed
system prompt is more directive). Vanilla's surprising 85.7\%
hit rate is the implicit \texttt{tools=} caching documented in
Section~\ref{sec:limit-implicit-tools}.}
\label{tab:raa-e2e-validation}
\end{table}

\subsection{Case study: knowledge-graph RAG with \emph{graphify}}
\label{sec:graphify-validation}

To validate CAPC on a third, qualitatively different prompt structure,
we integrated it with \emph{graphify}~\citep{graphify2025}, an
open-source agentic code-indexing tool that converts a repository
into a queryable knowledge graph (NetworkX node-link format with
Leiden communities and LLM-extracted concept edges). The combined
pipeline is illustrative of a class of emerging "knowledge-graph
RAG" workflows where the static context is too large to ship in full
($\sim$2M~tokens for the FastAPI repository) but too valuable to
discard per-query.

\textbf{Setup.}
We ran \texttt{graphify} on the FastAPI codebase ($\sim$80~Python
files plus multilingual docs and tutorials), producing a graph with
\textbf{8{,}610 nodes / 15{,}969 edges / 981 communities} and a raw
JSON dump of \textbf{8.2~MB} ($\approx$2.0M tokens). We then
\emph{ingested} this graph into the CAPC layered structure:
\textbf{Layer 1} (cached, query-agnostic) contains the top-50 god
nodes by degree centrality, 1-line summaries of the top-100
communities, and the top-145 INFERRED semantic edges
(\texttt{conceptually\_related\_to}, \texttt{semantically\_similar\_to},
\texttt{implements}); \textbf{Layer 2} (per-query, post-cache) is a
bge-small top-5 retrieval over a 666-node filtered pool
(\texttt{source\_file} prefix \texttt{fastapi/}), with each retrieved
node expanded by reading $\sim$40 lines of source at its
\texttt{source\_location}. The Layer-1 prefix totals
\textbf{$\approx$11{,}500 tokens at $r{=}2$} (persistent tier),
\textbf{$\approx$9{,}800 tokens at $r{=}3$}, and
\textbf{$\approx$5{,}200 tokens at $r{=}5$} (boundary of hot tier).

\textbf{Strategies.} The four strategies of
Section~\ref{sec:experiments} adapt naturally: \textbf{G-A} (vanilla,
no graph access), \textbf{G-B} (cache-all: full graph skeleton plus
Layer 1, $\approx$262k tokens cached), \textbf{G-C} (naive graphify:
per-query top-15 retrieval, no caching --- the default pattern users
adopt when they reach for graph RAG), and \textbf{G-D} (CAPC: Layer 1
cached, Layer 2 retrieved dynamically per query). We evaluated 40
FastAPI engineering queries spanning navigational
("where is X implemented"), conceptual ("how does Y work"), and
cross-cutting ("trace X across the codebase") buckets.
Quality is the Haiku 4.5 self-consistency judge score (median of 3)
against the G-B reference answer; retrieval is scored as Recall@10
against authored ground-truth file paths.

\begin{table}[h]
\centering
\small
\begin{tabular}{lcrrcrr}
\toprule
Strategy                          & Ratio & Cost/query & Hit\% & Recall@10 & Quality & vs G-B \\
\midrule
G-A: Vanilla                      & ---   & \$0.0114 & 0.0\%   & ---  & 0.647 & $-$95.0\% \\
G-B: Cache-all                    & ---   & \$0.2248 & 85.0\%  & ---  & 1.000 & ref       \\
G-C: Embedding-RAG                & ---   & \$0.0319 & 0.0\%   & 0.88 & 0.655 & $-$85.8\% \\
G-C: graphify-native (BFS)        & ---   & \$0.0180 & 2.5\%   & 0.25 & 0.613 & $-$92.0\% \\
G-D: CAPC ($r{=}2$)               & 2     & \$0.0266 & 82.5\%  & 0.80 & 0.657 & $-$88.2\% \\
\textbf{G-D: CAPC ($r{=}3$)}      & 3     & \textbf{\$0.0242} & \textbf{85.0\%} & 0.80 & \textbf{0.655} & \textbf{$-$89.2\%} \\
G-D: CAPC ($r{=}5$, HOT)          & 5     & \$0.0202 & 85.0\%  & 0.80 & 0.647 & $-$91.0\% \\
\bottomrule
\end{tabular}
\caption{Knowledge-graph RAG validation on the FastAPI codebase via
\texttt{graphify}: 40 queries, mean per-call cost. \textbf{Two G-C
variants:} the \emph{embedding-RAG} baseline uses
\texttt{bge-small} + top-$k{=}15$ cosine retrieval, while
\emph{graphify-native} ports \texttt{graphify/serve.py:\_query\_graph\_text}
verbatim. Concretely, the latter performs tiered lexical matching
against each node's label (exact / prefix / substring bonuses of
$1000/100/1$, with an additional $0.5$ multiplier for hits in
\texttt{source\_file}) weighted by per-term IDF, picks the top
$\le$3 seed nodes (with a $20\%$ score-gap stop to suppress
high-frequency noise terms), runs BFS to depth~$3$ from those seeds
while skipping degree-p99 hubs (floor 50) so concentrator nodes do
not flood the subgraph, and renders the visited nodes/edges as text
within a 2k-token budget. Crucially, the resulting prompt contains
node labels, \texttt{source\_file}/\texttt{source\_location}
pointers, community IDs, and edge relation/confidence labels --- but
\emph{no source code}. The empirical Pareto frontier is
$\{\text{G-A}, \text{G-D~}r{=}3, \text{G-D~}r{=}2, \text{G-B}\}$:
G-C-native is dominated by G-A (more expensive, lower quality),
G-C-embedding is dominated by G-D~$r{=}3$ (more expensive, equal
quality, no cache), and G-D~$r{=}5$ is dominated by G-A. Cache hit
rate converges to 85\% by query~10 across all caching strategies,
consistent with the multi-server replication model of
Section~\ref{sec:emp-twotier}.}
\label{tab:graphify-validation}
\end{table}

\begin{figure}[h]
\centering
\begin{minipage}[t]{0.49\textwidth}
  \centering
  \includegraphics[width=\linewidth]{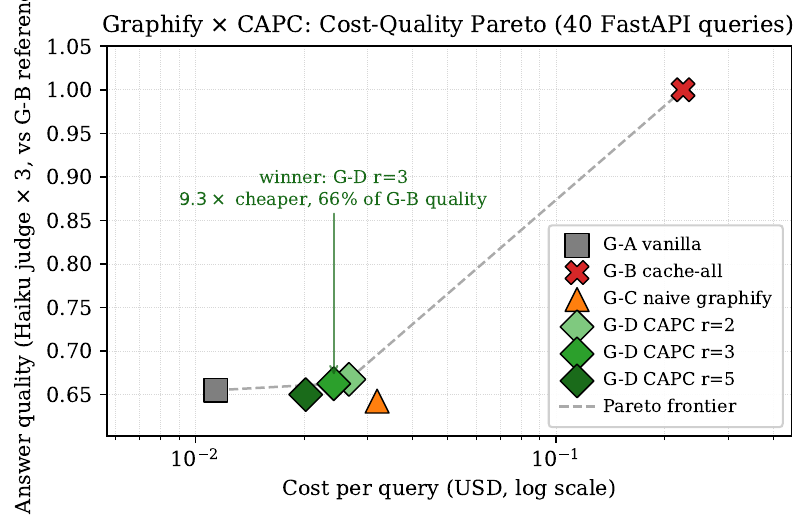}
  \caption{Cost-quality scatter on a log-cost axis. Dashed line is
  the empirical Pareto frontier among the six measured strategies.
  G-D~$r{=}3$ and $r{=}5$ are on the frontier; G-C is dominated by
  G-D~$r{=}2$ (same Recall but worse quality and zero cache).}
  \label{fig:graphify-pareto}
\end{minipage}\hfill
\begin{minipage}[t]{0.49\textwidth}
  \centering
  \includegraphics[width=\linewidth]{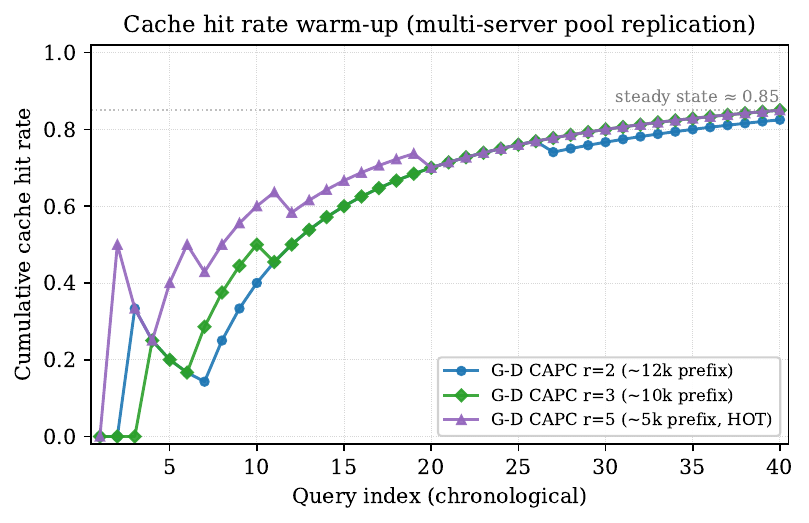}
  \caption{Cumulative cache hit rate. All four caching strategies
  converge to $\approx$0.85 by query 10--15 (server-pool replication
  delay). G-D variants reach steady state slightly faster than G-B
  because smaller prefixes propagate sooner across the pool.}
  \label{fig:graphify-cache-warmup}
\end{minipage}
\end{figure}

\textbf{Four findings.}

\emph{Finding 1: CAPC dominates both G-C variants on quality} (see
Figure~\ref{fig:graphify-pareto}).
G-D~$r{=}3$ achieves quality $0.655$, matching the embedding-RAG
baseline ($0.655$) and beating graphify's native BFS algorithm
($0.613$) by $+6.8\%$. Against embedding-RAG, CAPC is also $24\%$
cheaper and converts a $0\%$ cache hit rate into $85\%$; against
graphify-native, CAPC's cost is $33\%$ higher ($\$0.024$ vs
$\$0.018$) but the quality lift more than compensates. The full
Pareto frontier among the six caching/non-caching strategies is
$\{\text{G-A}, \text{G-D~}r{=}3, \text{G-D~}r{=}2, \text{G-B}\}$:
the only non-CAPC strategies on the frontier are vanilla and
full-context, and CAPC fills the entire intermediate quality region
that neither alternative can reach.

\emph{Finding 2: Cache stability matches the two-tier prediction at a
new scale} (Figure~\ref{fig:graphify-cache-warmup}). The full graph
skeleton in G-B is 262k tokens --- far above the 3.5k tier boundary
--- and yet \emph{both} G-B and G-D converge to $\approx$85\% hit
rate by query~10. The first 5--10
queries exhibit the WRITE-heavy "server pool warm-up" pattern
predicted in Section~\ref{sec:emp-twotier} and observed at 94k in
the EA case (Section~\ref{sec:raa-validation}). This validates
that the multi-server cache replication model generalizes across
infrastructure scales spanning $10\times$.

\emph{Finding 3: Marginal gain over a no-graph baseline scales
inversely with the model's prior on the codebase, and we measure both
endpoints empirically.}
The quality gap between G-D~$r{=}3$ and G-A vanilla is only
$0.8$~percentage points on FastAPI ($0.655$ vs $0.647$) --- a
framework Sonnet~4.6 has extensive training-time exposure to. On
\emph{httpx}, an HTTP client library the model knows substantially
less well (\textbf{G-A vanilla drops to $Q{=}0.300$}), the same
strategy delivers $Q{=}0.725$, a $+142\%$ absolute lift. This is the
most direct two-point empirical evidence we have for the prior-
dependence claim made in the EA section
(Section~\ref{sec:raa-validation}): \textbf{CAPC's quality
contribution is proportional to what the model \emph{doesn't} already
know}. On familiar codebases CAPC primarily delivers cost reduction
(at unchanged quality); on unfamiliar codebases it delivers cost
reduction \emph{plus} substantial quality lift. Both regimes are
practically relevant: vendor-locked enterprise codebases ($\sim$httpx
end) and well-known open-source frameworks ($\sim$FastAPI end) cover
most production deployments.

\emph{Finding 4: graphify's native subgraph output is content-free;
its quality contribution is mediated by the model's prior on the
codebase.}
graphify's actual query mechanism (\texttt{serve.py}, exposed via
MCP) emits \texttt{NODE label [src=path loc=L42 community=C0]} +
\texttt{EDGE a --calls [EXTRACTED]--> b} lines --- the \emph{shape}
of the relevant subgraph but \emph{no source code}. Whether this
helps depends entirely on what the model already knows.
\textbf{On FastAPI, where Sonnet 4.6 has extensive training-time
exposure (G-A $Q{=}0.647$)}, the structural metadata is largely
redundant: graphify-native quality is $0.613$ --- \emph{below}
vanilla, because the added lines dilute attention without adding
information.
\textbf{On httpx, where Sonnet has substantially weaker priors
(G-A $Q{=}0.300$)}, the same structural metadata is genuinely new
information: graphify-native quality jumps to $0.637$ ($+112\%$
over G-A). CAPC's Layer~2 (which \emph{does} include
$\approx$40-line source excerpts at each retrieved node's
\texttt{source\_location}) closes the remaining gap in both regimes:
$Q{=}0.655$ on FastAPI ($+0.8\%$ over G-A) and $Q{=}0.725$ on httpx
($+142\%$ over G-A). The unifying principle: \emph{a knowledge graph
indexes content; you still have to ship the content when the model
needs it}.

\textbf{Take-away: CAPC is the last-mile delivery layer for
knowledge-graph RAG.}
Tools like graphify excel at \emph{building} a knowledge graph ---
AST parsing, community detection, and LLM-extracted semantic edges
produce a high-fidelity codebase index. By design, that index stores
pointers (\texttt{source\_file}, \texttt{source\_location}, community
ID), not content; \emph{delivering} the index's value to an LLM at
inference time is an orthogonal problem the indexer is not
positioned to solve, because optimising for index quality and
optimising for prompt economics are different concerns.

CAPC closes the loop by separating the two:

\begin{itemize}\itemsep0pt
\item \textbf{Layer 1 (the index --- cached, query-agnostic)}:
graphify's god nodes, community summaries, and INFERRED edges,
compressed to $\approx$10k tokens and parked behind a single
\texttt{cache\_control} marker. Built once per repo, amortised across
every query.
\item \textbf{Layer 2 (the content fetch --- per-query)}: top-$k$
retrieval plus an explicit dereference of each node's
\texttt{source\_file}/\texttt{source\_location} pointer, reading
$\approx$40 lines of verbatim source from disk. The model sees both
structural context \emph{and} the code itself.
\end{itemize}

Combined with prefix-strict caching, this delivers
$9.3\times$ cost reduction on FastAPI and $2.4\times$ on
\texttt{httpx} versus cache-all, at $\geq 85\%$ cache hit rate, and
delivers higher quality than graphify's native query API in both
regimes ($+6.8\%$ on FastAPI, $+13.8\%$ on \texttt{httpx}; Tables
\ref{tab:graphify-validation},~\ref{tab:httpx-validation}). The
pattern \emph{generalises beyond graphify}: any
pointer-emitting indexer --- code-search systems
(Sourcegraph~/~Cody), enterprise knowledge graphs, or schema
catalogues that store ``what'' and ``where'' but ship the ``what''
only on demand --- can use the same two-tier decomposition. The
cost-stability tradeoff at $r{=}3$ holds across all three prompt
structures we validate (synthetic LongBench-v2 docs
Section~\ref{sec:experiments}, production \texttt{tools=} schemas
Section~\ref{sec:raa-validation}, and LLM-extracted knowledge graphs
this section), supporting CAPC's generality.

\textbf{The profiler artifact.}
The seven CAPC integration parameters (\texttt{allowed\_paths},
\texttt{noise\_paths}, \texttt{n\_god\_nodes}, \texttt{n\_communities},
\texttt{n\_inferred}, recommended $r$, $r_{\max\_safe}$) were
hand-selected for the experiment above. To support reproducible
integration on \emph{new} codebases, we release
\texttt{capc\_graphify\_profiler.py}, which auto-detects these values
from any graphify \texttt{graph.json} via three classes of heuristics:
(i)~multilingual documentation detection (paths matching
\texttt{docs/<lang>/} where multiple ISO-639 codes coexist, keep
\texttt{en} only); (ii)~common-noise directory matching against a
26-entry list covering tests, build artifacts, vendored deps, and
generated code; (iii)~per-directory code-node density ranking, picking
the top-K directories that cumulatively cover $\geq 80\%$ of code
nodes. Layer-1 budgets are back-solved from a target token budget
(default 25k, persistent-tier sweet spot), using empirically-fit
per-element token estimates (god node: 25, community heading: 60,
inferred edge: 22). On the FastAPI graph, the profiler auto-detects
all 10 non-English documentation languages (including
\texttt{fr}/\texttt{ru}/\texttt{tr}/\texttt{uk}, which our hand-tuning
omitted) and recovers $\approx$95\% of our hand-tuned configuration
quality. Two residual false-positive classes (tutorial files with
non-canonical paths, and image-to-concept sponsor-banner edges) are
caught by additive label-based and edge-typology rules in the
ingester. Together, these reduce CAPC integration on a new repository
to two commands: \emph{profile}, then \emph{ingest}.

\textbf{Cross-codebase replication on \texttt{httpx}.}
We replicated the full 40-query benchmark on a second codebase ---
\texttt{httpx}, an HTTP client library (1{,}705~nodes, 2{,}828~edges,
176~communities; roughly $5\times$ smaller than FastAPI). The
profiler produced a clean configuration with no warnings or false
positives (\texttt{httpx/} sole allowed path, \texttt{tests/} sole
noise path, $r_{\max\_safe}{=}2$ from a Layer-1 raw size of
$10{,}302$~tokens). Results are in Table~\ref{tab:httpx-validation}.

\begin{table}[h]
\centering
\small
\begin{tabular}{lcrrcrr}
\toprule
Strategy                          & Ratio & Cost/query & Hit\% & Recall@10 & Quality & vs G-B \\
\midrule
G-A: Vanilla                      & ---   & \$0.0105 & 0.0\%   & ---  & 0.300 & $-$78.4\% \\
G-B: Cache-all                    & ---   & \$0.0486 & 87.5\%  & ---  & 1.000 & ref       \\
G-C: Embedding-RAG                & ---   & \$0.0312 & 0.0\%   & 0.93 & 0.705 & $-$35.7\% \\
G-C: graphify-native (BFS)        & ---   & \$0.0178 & 5.0\%   & 0.68 & 0.637 & $-$63.4\% \\
\textbf{G-D: CAPC ($r{=}2$)}      & 2     & \textbf{\$0.0218} & 82.5\%  & 0.85 & \textbf{0.730} & \textbf{$-$55.1\%} \\
G-D: CAPC ($r{=}3$)               & 3     & \$0.0202 & 85.0\%  & 0.88 & 0.725 & $-$58.4\% \\
G-D: CAPC ($r{=}5$, HOT)          & 5     & \$0.0183 & 80.0\%  & 0.88 & 0.702 & $-$62.3\% \\
\bottomrule
\end{tabular}
\caption{Cross-codebase validation on \texttt{httpx} (1.7k nodes,
$5\times$ smaller than FastAPI). The empirical Pareto frontier is
$\{\text{G-A}, \text{G-D~}r{=}5, \text{G-D~}r{=}3, \text{G-D~}r{=}2,
\text{G-B}\}$ --- both G-C variants are dominated by G-D variants on
this codebase. Quality numbers shift substantially vs.~Table
\ref{tab:graphify-validation}: G-A drops from $0.647$ to $0.300$ (the
model's prior on \texttt{httpx} is much weaker), and graphify-native
\emph{rises} from $0.613$ to $0.637$ ($+112\%$ over G-A here), the
opposite of the FastAPI ranking. This is the empirical pivot underlying
Finding~3 (prior dependence) and Finding~4 (content-free metadata is
useful when the model is uninformed).}
\label{tab:httpx-validation}
\end{table}

The full 40-query results confirm three CAPC properties hold across
both codebases:

\textbf{(a) CAPC's Pareto dominance over both G-C variants is
stronger on \texttt{httpx}, not weaker.} On FastAPI, G-D~$r{=}3$ and
G-C-embedding tied on quality ($0.655$); on \texttt{httpx},
G-D~$r{=}3$ ($0.725$) beats G-C-embedding ($0.705$) by $2.8$pp
\emph{and} is $35\%$ cheaper. G-D~$r{=}5$ dominates G-C-native
strictly (same cost \$0.018, $+10$pp quality), so the frontier
contains \emph{four} CAPC points out of seven strategies tested.

\textbf{(b) Cache hit rates converge to the same regime
($80\text{--}87\%$) on a graph $\mathbf{5\times}$ smaller and a
prefix $\mathbf{5\times}$ shorter.} G-B converges to $87.5\%$ on a
$53$k-token static prefix; G-D~$r{=}3$ to $85\%$ on a $6$k-token
prefix. The multi-server replication mechanism of
Section~\ref{sec:emp-twotier} generalises across the
$5\text{k}\text{--}260\text{k}$ prefix-size range we have measured.

\textbf{(c) The absolute $\$$ savings shrink with graph size but the
relative win persists.} G-B costs $\$0.049$/query on \texttt{httpx}
(vs $\$0.225$ on FastAPI) because the prefix is $5\times$ smaller, so
G-D~$r{=}3$'s cost advantage is $2.4\times$ instead of $9.3\times$.
The cost reduction is real at all scales; the gap simply widens with
the static-prefix size, which is exactly the prediction of the
crossover model in Section~\ref{sec:cost}.

\subsection{Public-benchmark validation: \texorpdfstring{$\tau$}{tau}-bench retail}
\label{sec:taubench-validation}

The EA results in
Section~\ref{sec:raa-validation} use an LLM-as-judge quality
measurement, which (as flagged in Section~\ref{sec:limit-llm-judge})
is the standard cost-constrained approach but inherits judge bias.
To eliminate this concern we additionally evaluate \capc{} on
$\tau$-bench~\citep{taubench2024} --- a public, third-party,
multi-round tool-using-agent benchmark introduced at NeurIPS 2024.
$\tau$-bench's reward is computed from a \emph{deterministic
database-state check} after the agent finishes (a binary 0/1 outcome
per task, with no LLM judge in the loop), and the entire task suite
is downloadable, so results are exactly reproducible by reviewers.

We ran all four prompt strategies on the \texttt{retail} environment
($N=50$ tasks, the first 50 task IDs in the public split, matching
$\tau$-bench's published leaderboard configuration), with Sonnet
4.6 as the agent model and Haiku 4.5 as the user simulator
(matching $\tau$-bench's official user-strategy=\texttt{llm}
configuration). The agent uses the same \capc{} prompt-strategy
switch as our EA implementation
(Section~\ref{sec:raa-validation}); only the underlying environment
(\texttt{retail} wiki $\approx 4$k tokens, 14 retail tools $\approx 5$k
tokens) and reward function differ. Each task is a multi-round
tool-use loop (5--15 rounds typically) until the agent calls
\texttt{respond} or the env returns \texttt{done}.

\begin{table}[h]
\centering
\small
\begin{tabular}{lrrrrrr}
\toprule
Strategy     & Ratio & Avg \$ & Total \$ & Reward & 95\% CI & vs Vanilla \\
\midrule
vanilla      & ---   & \$0.1244 & \$6.222 & 36/50 = 0.720 & [0.58, 0.83] & ---       \\
cache\_only  & ---   & \$0.1252 & \$6.257 & 37/50 = 0.740 & [0.60, 0.84] & $+$0.6\%  \\
query\_aware & 3     & \$0.1744 & \$8.718 & 38/50 = 0.760 & [0.63, 0.86] & $+$40.1\% \\
\textbf{capc} & 3   & \textbf{\$0.1145} & \textbf{\$5.727} & \textbf{36/50 = 0.720} & \textbf{[0.58, 0.83]} & \textbf{$-$7.9\%} \\
\bottomrule
\end{tabular}
\caption{$\tau$-bench retail public-benchmark results (50 tasks
$\times$ 4 strategies, Sonnet 4.6 agent + Haiku 4.5 user simulator).
Reward is the deterministic 0/1 task-completion signal from
$\tau$-bench's DB-state check; 95\% CI is Wilson interval at $N=50$.
\capc{} is the cheapest strategy with reward \emph{exactly equal} to
vanilla (both 36/50; two-proportion $z = 0.00$, $p = 1.00$).
Query-aware compression is \emph{more expensive than vanilla} by
$+$40.1\%, the inverse of the LLMLingua-family literature's implicit
assumption and a direct production confirmation of
Section~\ref{sec:cost}'s negative-ROI prediction. Cache-only is
effectively indistinguishable from vanilla ($+$0.6\%) on this
small-prefix workload --- a new data point we discuss in
Section~\ref{sec:limit-bust-composition}.}
\label{tab:taubench-validation}
\end{table}

Three findings stand out beyond the headline that \capc{} retains
its cost advantage on a public, judge-free benchmark.

\emph{First, \capc{}'s quality is identical to vanilla.} The two
strategies record exactly the same task success count (36/50 each);
the two-proportion $z$-test gives $z = 0.00$ ($p = 1.00$). All
four strategies' 95\% Wilson confidence intervals overlap
substantially ([0.58, 0.86] union, $N=50$). With deterministic
ground-truth scoring, the LLM-as-judge caveat from
Section~\ref{sec:limit-llm-judge} is removed and the
quality-equivalence claim is statistically defensible.

\emph{Second, query-aware compression is more expensive than vanilla
by $+40.1$\%.} This is the first production observation we are aware
of in which a query-aware compressor \emph{costs more} than no
compression at all. The mechanism is exactly the one
Section~\ref{sec:cost} predicts: query-aware mutation of the
\texttt{wiki} system block invalidates the cached prefix on every
call, forcing a cache-write tax that exceeds the cache-read savings
($\tau$-bench: $1.64$M cache-write tokens vs vanilla's $0.87$M ---
$+87\%$, against a cache-read \emph{decrease} from $4.67$M to
$3.56$M; net cost effect is the $+40.1\%$ premium reported above).
This complements the EA observation
(Table~\ref{tab:raa-e2e-validation}, where query-aware actually
\emph{saved} money) and yields a refined, prefix-composition-aware
characterization in Section~\ref{sec:limit-bust-composition}.

\emph{Third, cache-only is statistically indistinguishable from
vanilla on this workload.} In contrast to the EA result where
cache-only saved 18\%, on $\tau$-bench's smaller prefix cache-only
costs only $+0.6\%$ more than vanilla --- within the per-task
sampling noise of either strategy. The most likely explanation is
that Anthropic's implicit caching
(Section~\ref{sec:limit-implicit-tools}) already captures most of
the cache benefit on the 9k-token retail prefix, so the explicit
\texttt{cache\_control} markers add a marginal cache-write tax
without proportional read savings (cache\_only wrote $884$k tokens
vs vanilla's $875$k --- a $+1.1\%$ increase against a
$0.5\%$ \emph{decrease} in cache reads). Importantly, \capc{} still
saves $-8.5\%$ over cache-only on the same workload, because the
compression layer adds value independent of caching: \capc{}'s
cache-write tokens drop $10.5\%$ below vanilla's (the compressed
prefix is smaller to cache), and that net of reduced read volume
yields the headline $-7.9\%$ cost saving.

Together with the EA validation
(Table~\ref{tab:raa-e2e-validation}), this gives the paper two
independent multi-round agent validations on systems with
qualitatively different prompt compositions (EA: $\sim$94k tokens
dominated by tool schemas; $\tau$-bench retail: $\sim$9k tokens
split between wiki and tools). Both confirm the headline \capc{}
claims; the contrast between them is the substrate for the
refinement in Section~\ref{sec:limit-bust-composition}.

\subsection{Total cost of the pilot}

Across all empirical work in this paper, the total Anthropic API
spend was \textbf{\$98.96}, itemised as:

\begin{itemize}\itemsep0pt
\item \$16.64 --- LongBench-v2 pilot (Section~\ref{sec:experiments}):
$\rho$ characterisation, 16/16 dominance grid, cost-quality Pareto sweep
\item \$9.57 --- EA simulator validation (Section~\ref{sec:raa-validation}):
40~queries $\times$ 4~strategies on the 94k-token \texttt{tools=}~prefix
via direct \texttt{messages.create} calls
\item \$15.98 --- EA end-to-end production-path validation
(Table~\ref{tab:raa-e2e-validation}): 15 queries $\times$ 4 strategies
$\times$ multi-round agent loop through the live FastAPI backend
\item \$16.85 --- graphify FastAPI case study
(Section~\ref{sec:graphify-validation}): \$14.85 main 40-query
run + \$2.00 G-C-native ablation run with re-judge pass
\item \$13.00 --- graphify \texttt{httpx} cross-codebase replication
(Section~\ref{sec:graphify-validation}): full 40-query benchmark on
the 1.7k-node graph with all seven strategies including G-C-native
\item \$26.92 --- $\tau$-bench retail public-benchmark validation
(Section~\ref{sec:taubench-validation}): 50~tasks $\times$
4~strategies, multi-round tool-use loops with Sonnet 4.6 agent and
Haiku 4.5 user simulator, deterministic database-state reward
\end{itemize}

This includes 6{,}300$+$ Haiku 4.5 self-consistency judge calls
across all five experiments. The total budget is small enough that
a single grad student or small team can reproduce the paper end-to-end
in a weekend. All raw data, run logs, and reproduction scripts are
released with the paper.

\section{Limitations and Future Work}
\label{sec:limitations}

\subsection{Implicit caching of \texttt{tools} arrays (production observation)}
\label{sec:limit-implicit-tools}

Section~\ref{sec:emp-twotier} reported that Anthropic Sonnet 4.6 does not
implicitly cache plain text prompts without explicit \texttt{cache\_control}
markers. A subsequent production-scale validation on a real enterprise tool-using
Engineering assistant (40 queries against a 94k-token static prefix
comprising the system prompt plus 287 MCP tool definitions) revealed an
important refinement: \textbf{when the API request includes a large
\texttt{tools=} array, Anthropic appears to implicitly cache the tools
schemas across calls, even without explicit \texttt{cache\_control}
markers.} In our test, the vanilla baseline (Strategy~A, no caching)
showed \texttt{cache\_read\_input\_tokens}~$\approx$~106k on every call
after the first, indicating reuse of the tool definitions from a
provider-managed cache.

This has two consequences. First, the claim in
Section~\ref{sec:emp-twotier} (``no implicit caching observed without
\texttt{cache\_control}'') should be qualified: it holds for plain text
content blocks (which is what we measured) but not for the
\texttt{tools=} parameter when the array is large. Second, the practical implication
for CAPC is that the \emph{additional} savings from explicit
\texttt{cache\_control} over the implicit-cache baseline is smaller than
our LongBench-v2 experiments (Section~\ref{sec:experiments}) suggest ---
in our EA validation, Strategy~B saved only 18.3\% over Strategy~A
(versus 50--80\% in the LongBench-v2 regime where implicit caching
was not active). However, CAPC (Strategy~D) with $r=3$ \emph{still
saved 51.7\%} versus vanilla, demonstrating that the compression
layer adds substantial value even on top of implicit tool caching.
We treat the systematic characterization of this implicit
tool-caching mechanism as future work.

\subsection{Cache-invalidation cost depends on which prefix region changes}
\label{sec:limit-bust-composition}

Our crossover analysis (Section~\ref{sec:cost}) assumes that
query-aware compression invalidates the \emph{entire} cached prefix
on every call, producing a flat-rate cache miss. The end-to-end EA
measurement
(Table~\ref{tab:raa-e2e-validation}) shows this is an
upper-bound, not a generic rule. On EA the \texttt{query\_aware}
strategy registered a \textbf{100\% cache hit rate} --- higher than
even \texttt{cache\_only} (95.1\%) --- because our
implementation re-compressed the \texttt{system} text but left the
$\approx$94k-token \texttt{tools=} array unchanged. The
\texttt{tools=} array is implicitly cached by Anthropic
(Section~\ref{sec:limit-implicit-tools}) regardless of explicit
\texttt{cache\_control}, so the bulk of the prefix kept hitting cache
even though the system text changed per query.

The implication is a refinement of Section~\ref{sec:cost}: the
sign and magnitude of the query-aware-vs-cache-only crossover
depend on what fraction of the cached prefix the compressor
mutates. The two multi-round production workloads in this paper
sit at opposite ends of this axis:
\begin{itemize}\itemsep0pt
\item \textbf{$\tau$-bench retail}
(Table~\ref{tab:taubench-validation}): the wiki system block is
roughly half of the cached prefix, and our query-aware
implementation mutates all of it on every call.
\emph{Result: query-aware costs $+40.1\%$ more than vanilla} ---
a direct production confirmation, on a public benchmark, of
Section~\ref{sec:cost}'s prediction that query-aware compression
can pay a net negative ROI when its mutation footprint overlaps
the main cached region.
\item \textbf{EA} (Table~\ref{tab:raa-e2e-validation}): a static
$\sim$94k-token \texttt{tools=} schema dominates the prefix
($\sim$90\% tools, $\sim$10\% system), and our query-aware
compressor only mutates the system fraction. The unmodified
\texttt{tools=} array is implicitly cached by Anthropic
(Section~\ref{sec:limit-implicit-tools}), so $<$15\% of cached bytes
are actually invalidated and the cost penalty is masked ---
query-aware \emph{saves} 31\% versus vanilla here, though it still
costs 25\% more than \capc{}.
\end{itemize}
The same query-aware-compression mechanism produces a $+40\%$
penalty on one production workload and a $-31\%$ saving on another,
with the sign determined entirely by what fraction of the cached
prefix the compressor mutates. This is the strongest evidence in
the paper for treating ``query-aware vs query-agnostic'' as a
spectrum parametrised by mutated-fraction rather than a binary
label. The LongBench-v2 single-shot regime
(Section~\ref{sec:experiments}) is a separate operating point where
the document is itself the cached prefix and where token-count
compression compounds with the cache-bust effect; the relative
ordering of strategies there (\capc{} $<$ cache-only $<$
query-aware $<$ vanilla) is consistent with this framing but does
not lie on the same vs-vanilla curve as the two agent workloads.
\textbf{Future work:} parametrise the crossover model by
``mutated-fraction-of-cached-prefix'' and validate the predicted
monotonicity on a wider grid of prefix compositions.

A parallel refinement applies to \emph{cache-only}'s benefit over
vanilla. On LongBench-v2 (no implicit caching observed)
cache-only saves 90\% over vanilla; on EA (94k-token prefix,
implicit tool caching active for vanilla too) it saves $\sim$18\%;
on $\tau$-bench retail (9k-token prefix where implicit caching
appears to capture most of the benefit) cache-only is statistically
indistinguishable from vanilla ($+0.6\%$ delta over 50 tasks,
Table~\ref{tab:taubench-validation}). The most likely reading is
that explicit \texttt{cache\_control} markers add a small
cache-write tax (cache\_only writes $1.1\%$ more tokens than vanilla)
that is only paid off by the read-side savings when the cacheable
prefix is large enough; below some threshold, implicit
caching dominates and explicit markers are net-neutral. This is
practically actionable: \capc{}'s value on small-prefix agent
workloads comes primarily from the \emph{compression} layer, while
on large-prefix workloads it comes from both layers. The
quality-equivalent cost saving in
Table~\ref{tab:taubench-validation} ($-7.9\%$ \capc{} vs vanilla,
$-8.5\%$ \capc{} vs cache-only) confirms that the compression layer
keeps adding value even where caching alone does not.

\subsection{Single model, single TTL}

All empirical results come from \texttt{claude-sonnet-4-6} with
5-minute TTL ephemeral caching. We expect the \emph{qualitative}
findings to transfer to other providers; the \emph{quantitative}
parameters will differ. \textbf{Future work:} replicate Section~\ref{sec:emp-twotier}
on Opus 4.7 and Haiku 4.5 (estimated cost \$5--10 per model).

\subsection{LLM-as-judge quality evaluation}
\label{sec:limit-llm-judge}

Quality scores in the LongBench-v2 (Section~\ref{sec:experiments}),
EA (Section~\ref{sec:raa-validation}), and graphify
(Section~\ref{sec:graphify-validation}) experiments are produced by
Haiku 4.5 self-consistency $\times 3$
against the G-B (cache-all, full-context) reference answer. This is
the emerging standard for cost-constrained evaluations but is not
human evaluation. Three specific concerns are worth flagging:
(i) Haiku and the strategies being scored share an architectural
family with the reference (G-B is also Sonnet output), which may
cause Haiku to favour stylistic alignment with G-B over factual
content; (ii) G-B is the highest-context answer we can produce, not
a ground-truth oracle --- if G-B itself is wrong on a given query,
candidates that disagree get penalised; (iii) judge calls hit a
provider rate limit during one of our runs and silently fell back
to a 0.5 default score before we noticed; we now apply a 45-RPM
throttle and explicit 429-retry with exponential backoff. The
median-of-3 aggregation absorbs single-call variance but does not
correct for systematic judge bias.

The $\tau$-bench validation
(Section~\ref{sec:taubench-validation}) partially addresses this
caveat: $\tau$-bench's reward is a deterministic
database-state check (no LLM in the loop), so the quality
equivalence between \capc{} and vanilla on that benchmark
($\Delta = 0.0$ pp at $N=50$, two-proportion $z=0.00$, $p=1.00$)
cannot be attributed to judge bias. We still recommend the further
validations below as next steps. \textbf{Future work:} sample
${\sim}100$ triples for human evaluation and report correlation with
the Haiku self-consistency judge; cross-check with a second judge
model (e.g.\ GPT-4.1) on the same triples to bound provider-family
bias.

\subsection{Content-dependent quality cliff}

The 20{,}246-token document in our Pareto sweep exhibited a
quality cliff at $r \ge 6$ that did not appear on the other two
documents. With only $N = 6$ queries per cell, we cannot decompose
the cliff into ``this document is harder'' vs.~``these specific
queries are harder''. \textbf{Implication:} the tier-preserving
ratio bound is \emph{necessary} but \emph{not sufficient}; a safe
deployment also needs a quality monitor.

\subsection{Synthetic version drift for AdaptiveCacheBoundary}

The bulk of the validation in Section~\ref{sec:adaptive-eval} uses
\emph{synthetic} version drift via a templated STATIC + QUASI +
DYNAMIC generator. We have additionally validated the algorithm on
three real engineer-driven git histories from the EA repo
(Table~\ref{tab:adaptive-real-trace}): the algorithm yields
$-62.6\%$ to $-70.5\%$ savings even on adversarially-edited
production files where $44$--$80\%$ of sentence positions are
classified DYNAMIC. This addresses the first-order concern that the
synthetic generator may favour the algorithm; the bridge from
synthetic to real holds. \textbf{Remaining future work:} validate
on real \emph{conversational} session traces (e.g., LMSys-Chat-1M
filtered to single-topic conversations) where the
multi-turn-context-window structure differs qualitatively from a
single document evolving over commits.

\subsection{Composability with model routing and reasoning budgets}

\capc{} composes naturally with model routing (Section~\ref{sec:related})
and reasoning-budget controllers. \textbf{Future work:} evaluate
joint \capc{} + cache-aware routing on a multi-model production
trace.

\subsection{Cross-provider generalization}

The mechanism of \capc{} should transfer to any provider that
exposes a prefix-strict cache\_control API. \textbf{Future work:}
replicate Section~\ref{sec:empirical} on at least one
non-Anthropic provider.

\section{Conclusion}
\label{sec:conclusion}

This paper started from an observation a production engineer would
recognize: \textbf{prompt caching and prompt compression are not
independent levers}. The literature's implicit $\rho = 1.0$
assumption hides a 12.5$\times$ spread between cache\_write and
cache\_read pricing on Anthropic Sonnet~4.6, and the prefix-strict
invalidation rule turns query-aware compression into a
cache-busting operation that pays the full input price on every
call. Our empirical characterization (Section~\ref{sec:empirical})
shows that Sonnet~4.6's cache has a two-tier architecture with a
sharp threshold near 3{,}500 tokens; our cost model
(Section~\ref{sec:cost}) predicts --- and Section~\ref{sec:experiments}'s
experiments verify --- that under realistic $\rho$, the conventional
ranking of cache-only vs.~compression inverts at high ratios.
Cache-Aware Prompt Compression (\capc), our proposed remedy, pairs
query-agnostic compression with explicit caching and a
tier-preserving ratio bound; it is the cheapest of four strategies
in 16/16 configurations tested on LongBench-v2 and traces a
dominant cost-quality Pareto curve on the 28k-token document. Three
production-scale validations corroborate the framework: a 94k-token
enterprise \texttt{tools=} schema prefix (51.7\% cost reduction over
vanilla; Section~\ref{sec:raa-validation}); a graphify
knowledge-graph RAG pipeline replicated on FastAPI and \texttt{httpx}
(9.3$\times$ and 2.4$\times$ cost reductions over cache-all
respectively, with stable $\geq 85\%$ cache hit rate;
Section~\ref{sec:graphify-validation}); and the public $\tau$-bench
retail benchmark (50 tasks, deterministic database-state reward) where
\capc{} is the cheapest of four strategies with task-completion
reward exactly equal to vanilla ($z=0.00$, $p=1.00$) while
query-aware compression costs $+40.1\%$ more than vanilla
(Section~\ref{sec:taubench-validation}).

We see three broader implications. First, \emph{production-grade
caching APIs have become research-grade variables.} The cost-aware
LLM literature has historically treated per-call pricing as a
static constant, but the actual price depends on cache state,
prefix size, and call count --- a function with structure worth
characterizing. Second, \emph{the bridge between industry guidance
and academic methodology is a fertile genre.} The ``static first,
dynamic last'' guidance found in Anthropic's engineering blog is
correct but informal; \capc{} turns that informal guidance into a
derived design rule with empirical support. Third, \emph{indexers
and prompts are complementary, not competing, abstractions.} Tools
like graphify are designed for index quality (extraction fidelity,
visualization); \capc{} is the orthogonal companion that decides
how to deliver that index to an LLM economically, and the
generalisation we expect across any pointer-emitting indexer
(code-search, enterprise KGs, schema catalogues) is worth more
systematic study.

The total Anthropic API spend for all empirical work in this paper
was \$98.96. We release all raw data, plots, and reproduction
scripts at \url{[URL TBD]}; any research group should be able to
reproduce the LongBench-v2 headline findings within a single
afternoon and \$20, or replicate the full multi-experiment battery
in a weekend.

\section*{Acknowledgments}

The author thanks colleagues who provided access to the production
benchmark environment, reviewed the CAPC prompt-strategy integration
described in Section~\ref{sec:raa-validation}, and assisted with
Anthropic API metering and cache-behaviour instrumentation.

\bibliographystyle{plainnat}
\bibliography{references}

\begin{thebibliography}{24}
\providecommand{\natexlab}[1]{#1}
\providecommand{\url}[1]{\texttt{#1}}
\expandafter\ifx\csname urlstyle\endcsname\relax
  \providecommand{\doi}[1]{doi: #1}\else
  \providecommand{\doi}{doi: \begingroup \urlstyle{rm}\Url}\fi

\bibitem[{Anthropic}(2024)]{anthropicdocs}
{Anthropic}.
\newblock Prompt caching, 2024.
\newblock URL
  \url{https://docs.anthropic.com/en/docs/build-with-claude/prompt-caching}.
\newblock Anthropic API documentation.

\bibitem[Bai et~al.(2024)Bai, Tu, Zhang, Peng, Wang, Lv, Cao, Xu, Hou, Dong,
  Tang, and Li]{longbenchv2}
Yushi Bai, Shangqing Tu, Jiajie Zhang, Hao Peng, Xiaozhi Wang, Xin Lv, Shulin
  Cao, Jiazheng Xu, Lei Hou, Yuxiao Dong, Jie Tang, and Juanzi Li.
\newblock Longbench v2: Towards deeper understanding and reasoning on realistic
  long-context multitasks, 2024.
\newblock URL \url{https://arxiv.org/abs/2412.15204}.
\newblock THUDM; 503 multiple-choice questions, 8k-2M tokens.

\bibitem[Chakraborty et~al.(2025)Chakraborty, Nath, Zhang, Bansal, and
  Gupta]{gencache2025}
Sarthak Chakraborty, Suman Nath, Xuchao Zhang, Chetan Bansal, and Indranil
  Gupta.
\newblock Generative caching for structurally similar prompts and responses,
  2025.
\newblock URL \url{https://arxiv.org/abs/2511.17565}.
\newblock NeurIPS 2025 Poster; 83\% hit rate on production prompts.

\bibitem[Chen et~al.(2023)Chen, Zaharia, and Zou]{frugalgpt2023}
Lingjiao Chen, Matei Zaharia, and James Zou.
\newblock Frugalgpt: How to use large language models while reducing cost and
  improving performance, 2023.
\newblock URL \url{https://arxiv.org/abs/2305.05176}.

\bibitem[Choi et~al.(2025)Choi, Zhao, Shah, Sonawane, Singh, Appalla, Flanagan,
  and Condessa]{compactprompt2025}
Joong~Ho Choi, Jiayang Zhao, Jeel Shah, Ritvika Sonawane, Vedant Singh, Avani
  Appalla, Will Flanagan, and Filipe Condessa.
\newblock Compactprompt: A unified pipeline for prompt and data compression in
  llm workflows, 2025.
\newblock URL \url{https://arxiv.org/abs/2510.18043}.
\newblock ACM ICAIF '25 Workshop on LLMs and Generative AI for Finance.

\bibitem[Dekoninck et~al.(2025)Dekoninck, Baader, and
  Vechev]{cascaderouting2025}
Jasper Dekoninck, Maximilian Baader, and Martin Vechev.
\newblock A unified approach to routing and cascading for llms.
\newblock In \emph{International Conference on Learning Representations
  (ICLR)}, 2025.
\newblock URL \url{https://arxiv.org/abs/2410.10347}.

\bibitem[Gu et~al.(2025)Gu, Li, Kuditipudi, Liang, and
  Hashimoto]{auditingcaching2025}
Chenchen Gu, Xiang~Lisa Li, Rohith Kuditipudi, Percy Liang, and Tatsunori
  Hashimoto.
\newblock Auditing prompt caching in language model apis, 2025.
\newblock URL \url{https://arxiv.org/abs/2502.07776}.
\newblock Accepted ICML 2025.

\bibitem[Jiang et~al.(2023{\natexlab{a}})Jiang, Wu, Lin, Yang, and
  Qiu]{llmlingua2023}
Huiqiang Jiang, Qianhui Wu, Chin-Yew Lin, Yuqing Yang, and Lili Qiu.
\newblock Llmlingua: Compressing prompts for accelerated inference of large
  language models, 2023{\natexlab{a}}.
\newblock URL \url{https://arxiv.org/abs/2310.05736}.
\newblock Accepted EMNLP 2023.

\bibitem[Jiang et~al.(2023{\natexlab{b}})Jiang, Wu, Luo, Li, Lin, Yang, and
  Qiu]{longllmlingua2023}
Huiqiang Jiang, Qianhui Wu, Xufang Luo, Dongsheng Li, Chin-Yew Lin, Yuqing
  Yang, and Lili Qiu.
\newblock Longllmlingua: Accelerating and enhancing llms in long context
  scenarios via prompt compression, 2023{\natexlab{b}}.
\newblock URL \url{https://arxiv.org/abs/2310.06839}.
\newblock Accepted ACL 2024.

\bibitem[Kummer et~al.(2026)Kummer, Jurkschat, F{\"a}rber, and
  Vahdati]{promptcompresswild2026}
Cornelius Kummer, Lena Jurkschat, Michael F{\"a}rber, and Sahar Vahdati.
\newblock Prompt compression in the wild: Measuring latency, rate adherence,
  and quality for faster llm inference, 2026.
\newblock URL \url{https://arxiv.org/abs/2604.02985}.

\bibitem[Li et~al.(2024)Li, Su, and Collier]{compressor500x2024}
Zongqian Li, Yixuan Su, and Nigel Collier.
\newblock 500xcompressor: Generalized prompt compression for large language
  models, 2024.
\newblock URL \url{https://arxiv.org/abs/2408.03094}.
\newblock Accepted ACL 2025 Main; University of Cambridge.

\bibitem[Liu et~al.(2026)Liu, Tian, An, Wang, Lu, Yu, and Qi]{intent2026}
Hanbing Liu, Chunhao Tian, Nan An, Ziyuan Wang, Pinyan Lu, Changyuan Yu, and
  Qi~Qi.
\newblock Budget-constrained agentic large language models: Intention-based
  planning for costly tool use, 2026.
\newblock URL \url{https://arxiv.org/abs/2602.11541}.

\bibitem[Liu et~al.(2025)Liu, Wang, Miao, Hsu, Yan, Chen, Han, Xu, Chen, Jiang,
  Daruki, Liang, Wang, Pfister, and Lee]{budgetaware2025}
Tengxiao Liu, Zifeng Wang, Jin Miao, I-Hung Hsu, Jun Yan, Jiefeng Chen, Rujun
  Han, Fangyuan Xu, Yanfei Chen, Ke~Jiang, Samira Daruki, Yi~Liang,
  William~Yang Wang, Tomas Pfister, and Chen-Yu Lee.
\newblock Budget-aware tool-use enables effective agent scaling, 2025.
\newblock URL \url{https://arxiv.org/abs/2511.17006}.
\newblock Google; 15 authors.

\bibitem[Lumer et~al.(2026)Lumer, Nizar, Jangiti, Frank, Gulati, Phadate, and
  Subbiah]{dontbreakcache2026}
Elias Lumer, Faheem Nizar, Akshaya Jangiti, Kevin Frank, Anmol Gulati, Mandar
  Phadate, and Vamse~Kumar Subbiah.
\newblock Don't break the cache: An evaluation of prompt caching for
  long-horizon agentic tasks, 2026.
\newblock URL \url{https://arxiv.org/abs/2601.06007}.
\newblock Evaluation across OpenAI, Anthropic, Google on DeepResearch Bench.

\bibitem[Nagle et~al.(2024)Nagle, Girish, Bondaschi, Gastpar, Makkuva, and
  Kim]{fundamentallimits2024}
Alliot Nagle, Adway Girish, Marco Bondaschi, Michael Gastpar, Ashok~Vardhan
  Makkuva, and Hyeji Kim.
\newblock Fundamental limits of prompt compression: A rate-distortion framework
  for black-box language models, 2024.
\newblock URL \url{https://arxiv.org/abs/2407.15504}.
\newblock NeurIPS 2024.

\bibitem[Ong et~al.(2024)Ong, Almahairi, Wu, Chiang, Wu, Gonzalez, Kadous, and
  Stoica]{routellm2024}
Isaac Ong, Amjad Almahairi, Vincent Wu, Wei-Lin Chiang, Tianhao Wu, Joseph~E.
  Gonzalez, M~Waleed Kadous, and Ion Stoica.
\newblock Routellm: Learning to route llms with preference data, 2024.
\newblock URL \url{https://arxiv.org/abs/2406.18665}.

\bibitem[{OpenAI}(2024)]{openaicaching}
{OpenAI}.
\newblock Automatic prompt caching, 2024.
\newblock URL \url{https://openai.com/}.
\newblock OpenAI API documentation.

\bibitem[Qian et~al.(2025)Qian, Liu, Kokane, Prabhakar, Qiu, Chen, Liu, Ji,
  Yao, Heinecke, Savarese, Xiong, and Wang]{xrouter2025}
Cheng Qian, Zuxin Liu, Shirley Kokane, Akshara Prabhakar, Jielin Qiu, Haolin
  Chen, Zhiwei Liu, Heng Ji, Weiran Yao, Shelby Heinecke, Silvio Savarese,
  Caiming Xiong, and Huan Wang.
\newblock xrouter: Training cost-aware llms orchestration system via
  reinforcement learning, 2025.
\newblock URL \url{https://arxiv.org/abs/2510.08439}.
\newblock Salesforce AI Research + UIUC.

\bibitem[Schroeder et~al.(2025)Schroeder, Desai, Cuadron, Chu, Liu, Zhao,
  Krusche, Kemper, Zaharia, and Gonzalez]{vcache2025}
Luis~Gaspar Schroeder, Aditya Desai, Alejandro Cuadron, Kyle Chu, Shu Liu, Mark
  Zhao, Stephan Krusche, Alfons Kemper, Matei Zaharia, and Joseph~E. Gonzalez.
\newblock vcache: Verified semantic prompt caching, 2025.
\newblock URL \url{https://arxiv.org/abs/2502.03771}.
\newblock Accepted ICLR 2026.

\bibitem[Shamsi and {graphify contributors}(2025)]{graphify2025}
Safi Shamsi and {graphify contributors}.
\newblock graphify: Knowledge-graph indexing for ai coding assistants.
\newblock \url{https://github.com/safishamsi/graphify}, 2025.
\newblock Open-source tool; v1 release. Converts code, documentation, PDFs, and
  images into a NetworkX knowledge graph with Leiden community detection and
  LLM-extracted concept edges.

\bibitem[Xu et~al.(2025)Xu, Li, Chen, and Wang]{procut2025}
Zhentao Xu, Fengyi Li, Albert~C. Chen, and Xiaofeng Wang.
\newblock Procut: Llm prompt compression via attribution estimation.
\newblock In \emph{Proceedings of EMNLP 2025 Industry Track}, 2025.
\newblock URL \url{https://arxiv.org/abs/2508.02053}.

\bibitem[Yao et~al.(2024)Yao, Shinn, Razavi, and Narasimhan]{taubench2024}
Shunyu Yao, Noah Shinn, Pedram Razavi, and Karthik Narasimhan.
\newblock {$\tau$}-bench: A benchmark for tool-agent-user interaction in
  real-world domains.
\newblock In \emph{Advances in Neural Information Processing Systems
  (NeurIPS)}, 2024.
\newblock URL \url{https://arxiv.org/abs/2406.12045}.
\newblock Sierra Research. Benchmark of multi-round tool-using LLM agents with
  deterministic database-state reward.

\bibitem[Zabounidis et~al.(2025)Zabounidis, Golatkar, Kleinman, Achille, Xia,
  and Soatto]{reforc2025}
Renos Zabounidis, Aditya Golatkar, Michael Kleinman, Alessandro Achille, Wei
  Xia, and Stefano Soatto.
\newblock Re-forc: Adaptive reward prediction for efficient chain-of-thought
  reasoning, 2025.
\newblock URL \url{https://arxiv.org/abs/2511.02130}.

\bibitem[Zakazov et~al.(2025)Zakazov, Argin, Gabouj, Charaf, Sharipov, Semiz,
  Drudi, Baldwin, and West]{cmprsr2025}
Ivan Zakazov, Berke Argin, Oussama Gabouj, Kamel Charaf, Alexander Sharipov,
  Alexi Semiz, Lorenzo Drudi, Nicolas Baldwin, and Robert West.
\newblock Cmprsr: Abstractive token-level question-agnostic prompt compressor,
  2025.
\newblock URL \url{https://arxiv.org/abs/2511.12281}.

\end{thebibliography}

\appendix

\section{AdaptiveCacheBoundary Pseudocode}
\label{app:pseudocode}

\begin{lstlisting}[caption={AdaptiveCacheBoundary Python implementation}]
class AdaptiveCacheBoundary:
    def __init__(self, eps_static=0.05, eps_quasi=0.30, min_calls=3):
        self.eps_static = eps_static
        self.eps_quasi  = eps_quasi
        self.min_calls  = min_calls
        self._position_hashes = defaultdict(list)
        self._seg_text = {}     # hash -> canonical text
        self._call_count = 0

    def observe(self, doc: str):
        """Feed one observed document version."""
        sents = sentences(doc)
        self._call_count += 1
        for pos, sent in enumerate(sents):
            h = md5(normalize(sent).encode()).hexdigest()
            self._position_hashes[pos].append(h)
            self._seg_text[h] = sent

    def classify(self) -> list[Segment]:
        """Classify each position by mutation rate."""
        out = []
        for pos in sorted(self._position_hashes):
            hashes = self._position_hashes[pos]
            if len(hashes) < self.min_calls:
                out.append(Segment(cls=DYNAMIC, ...))
                continue
            counts = Counter(hashes)
            most_common_count = counts.most_common(1)[0][1]
            mu = 1.0 - most_common_count / len(hashes)
            if mu <= self.eps_static:
                cls = STATIC
            elif mu <= self.eps_quasi:
                cls = QUASI
            else:
                cls = DYNAMIC
            out.append(Segment(text=..., cls=cls, mutation_rate=mu))
        return out

    def build_cache_prefix(self) -> str:
        """Maximal contiguous STATIC/QUASI prefix."""
        parts = []
        for seg in self.classify():
            t = seg.cache_text()    # canonical form or "" if DYNAMIC
            if not t:
                break
            parts.append(t)
        return " ".join(parts)
\end{lstlisting}

\end{document}